\pdfoutput=1

\documentclass[11pt]{article}

\usepackage[]{coling}

\usepackage{times}
\usepackage{latexsym}

\usepackage[T1]{fontenc}

\usepackage[utf8]{inputenc}

\usepackage{microtype}

\usepackage{inconsolata}

\usepackage{latexsym}
\usepackage{amssymb}
\usepackage{amsmath}
\usepackage{amsthm}
\usepackage{booktabs}
\usepackage{enumitem}
\usepackage{graphicx}
\usepackage{color}
\usepackage{microtype}
\usepackage{mathtools}  
\usepackage{subcaption}
\usepackage{xcolor, colortbl}
\usepackage{booktabs}
\usepackage{algorithm}
\usepackage{algorithmic}
\usepackage{epsfig}
\usepackage{graphicx}
\usepackage{amssymb}
\usepackage{epsfig}
\usepackage{amsmath, mdframed}
\usepackage{multirow}
\usepackage{makecell}

\newcolumntype{P}[1]{>{\centering\arraybackslash}p{#1}}
\usepackage{color,soulutf8}
\usepackage[export]{adjustbox}

\title {Enhancing Zero-shot Chain of Thought Prompting via Uncertainty-Guided Strategy Selection}

\author{ Shanu Kumar \quad Saish Mendke \quad Karody Lubna Abdul Rahman\\
\textbf{Santosh Kurasa \quad Parag Agrawal\quad Sandipan Dandapat}\\
Microsoft Corporation, India \\
{\tt \small \{shankum,saishmendke,lubnakarody,skurasa,paragag,sadandap\}@microsoft.com} }

\begin{document}
\maketitle
\begin{abstract}
Chain-of-thought (CoT) prompting has significantly enhanced the capability of large language models (LLMs) by structuring their reasoning processes. However, existing methods face critical limitations: handcrafted demonstrations require extensive human expertise, while trigger phrases are prone to inaccuracies. In this paper, we propose the \underline{Ze}ro-shot \underline{U}ncertainty-based \underline{S}election (\textit{ZEUS}) method, a novel approach that improves CoT prompting by utilizing uncertainty estimates to select effective demonstrations without needing access to model parameters. Unlike traditional methods, \textit{ZEUS} offers high sensitivity in distinguishing between helpful and ineffective questions, ensuring more precise and reliable selection. Our extensive evaluation shows that \textit{ZEUS} consistently outperforms existing CoT strategies across four challenging reasoning benchmarks, demonstrating its robustness and scalability. 
\end{abstract}

\section{Introduction}
Large Language Models (LLMs) have achieved remarkable performance in a wide range of natural language processing tasks~\cite{NEURIPS2020_1457c0d6, touvron2023llama, thoppilan2022lamda}. However, they often struggle with tasks that require complex reasoning~\cite{rae2021scaling, liang2022holistic}. The "chain-of-thought" (CoT) prompting technique~\cite{wei2022chain, feng2024towards} has been proposed to address this limitation by generating intermediate rationales ($r$) along with the final answer ($a$) for a given question ($q$). In this context, few-shot in-context examples, referred to as demonstrations $D = {(q_j, r_j, a_j)}_{j=1}^{k}$, consist of $k$ example questions $q_j$, manually crafted rationales $r_j$, and answers $a_j$. This approach, known as \textit{Manual-CoT}, relies on handcrafted rationales to guide the model.

Building on \textit{Manual-CoT}, \textit{Zero-Shot-CoT}~\cite{NEURIPS2022_8bb0d291} presents a novel prompting method where LLMs generate rationales using a trigger phrase $t$ (e.g., \textit{"Let's think step by step"}) appended to the input question $q$, without requiring manually crafted demonstrations. While Zero-Shot-CoT is cost-effective, its performance often falls short compared to \textit{Manual-CoT} due to the absence of effective demonstrations.

Crafting rationales manually is typically labor-intensive and time-consuming, particularly for tasks demanding intricate reasoning. To mitigate this, Auto-CoT~\cite{zhang2022automatic} combines Manual-CoT and Zero-Shot-CoT, thereby reducing the performance gap while minimizing manual effort. Auto-CoT employs self-supervised learning on a set of unlabeled questions $Q = \{{q_j}\}_{j=1}^{m}$ to generate rationales and answers. Demonstrations are created by clustering $Q$ into $k$ groups and selecting a representative question, rationale, and answer from each cluster. This clustering approach aims to maintain diversity in the demonstrations, which can help mitigate the impact of any errors in the generated rationales.

In this work, we seek to enhance the creation of demonstrations that improve LLM performance solely using unlabeled questions $Q$ without any rationale and answer. The selection process of examples $q_j$ in demonstrations $D$ has been to significantly influence LLM performance~\cite{wan-etal-2023-better}, and generating consistent rationales~\cite{wang2022self} is crucial. Recent CoT prompting methods~\cite{diao-etal-2024-active, bayer2024activellm} have utilized Active Learning (AL)~\cite{fu2013survey, settles2008analysis, rotman2022multi, kumar2022diversity} to identify examples for human annotation, showing that annotating the most uncertain examples yields the best performance. Drawing on these principles, we propose several selection strategies based on the uncertainty of unlabeled questions.

To estimate uncertainty, we adopt perturbation-based methods~\cite{ribeiro2020beyond, kuhn2023semantic, gao2024spuq, tomani2024uncertainty}, which operate on the principle that incorrect predictions can be detected by resampling rationales through perturbations, such as temperature adjustments. If the LLM is confident in its prediction, perturbations are unlikely to affect the outcome. However, if the LLM's prediction is uncertain, different perturbations can lead to varied responses. Our initial experiments reveal that while temperature-based perturbation estimates are well-calibrated, they lack sufficient sensitivity.\footnote{Sensitivity is a measure of the degree of change in accuracy by a unit change in the confidence score.} To address this, we propose a robust method for estimating uncertainty that exhibits near-ideal linearity with accuracy.

Our primary contributions are threefold: i) We present \textit{ZEUS},\footnote{We have uploaded code and datasets for reproducibility.} a method for estimating LLM uncertainty that is both well-calibrated and sensitive. ii) We leverage these uncertainty estimates to guide the selection of most informative demonstrations and show that these strategies outperform existing prompting methods across four challenging reasoning tasks. iii) We demonstrate that the performance of \textit{ZEUS} correlates strongly with few-shot uncertainty estimates on the unlabeled set, providing actionable recommendations for creating effective demonstrations.

\section{Related Work}
Chain-of-Thought (CoT) prompting has significantly influenced various advanced techniques designed to enhance reasoning capabilities. These include Tree of Thoughts~\cite{yao2023tree}, Role Play~\cite{kong-etal-2024-better}, and Collaborative Prompting~\cite{zhu-etal-2023-solving, yin-etal-2023-exchange, liang2023encouraging, wang2023unleashing}, each building on the CoT methodology to improve model performance in complex reasoning tasks. Concurrently, Active Learning (AL)-based methods have gained traction in few-shot prompting scenarios. \citet{diao2023active} enhance CoT prompting within an AL framework by actively selecting questions based on an uncertainty metric and manually constructing demonstrations. \citet{shum-etal-2023-automatic} work with labeled questions devoid of rationales, generating rationales through pruning and using an AL-inspired variance-reduced policy gradient strategy to select the most informative examples. Similarly, \citet{bayer2024activellm} apply uncertainty-based AL methods to identify the most valuable questions for annotation. Unlike these studies, our work addresses a more challenging scenario where neither human-annotated labels nor rationales are available.

Our method relies on accurate uncertainty estimation due to the lack of human supervision. Estimating uncertainty is a well-explored challenge, with methods ranging from Bayesian approaches and ensemble methods to more recent perturbation-based techniques~\cite{hendrycks2016baseline, gal2016dropout, NIPS2017_9ef2ed4b, pmlr-v70-guo17a, pmlr-v119-van-amersfoort20a, ovadia2019can}. Perturbation-based methods, which include techniques like temperature adjustments and question rephrasing, have shown promise in recent studies~\cite{gao2024spuq}. These methods, while effective, may not be universally applicable to LLMs due to their generative nature~\cite{vashurin2024benchmarking}. Other recent work has explored uncertainty estimation for specific tasks, such as hallucination detection in LLMs~\cite{kuhn2023semantic, tomani2024uncertainty}. We extend these perturbation techniques by enhancing their sensitivity to capture finer distinctions between questions, thereby improving uncertainty estimation in LLMs.

\begin{figure*}[htb]
    \centering
    \includegraphics[scale=0.16]{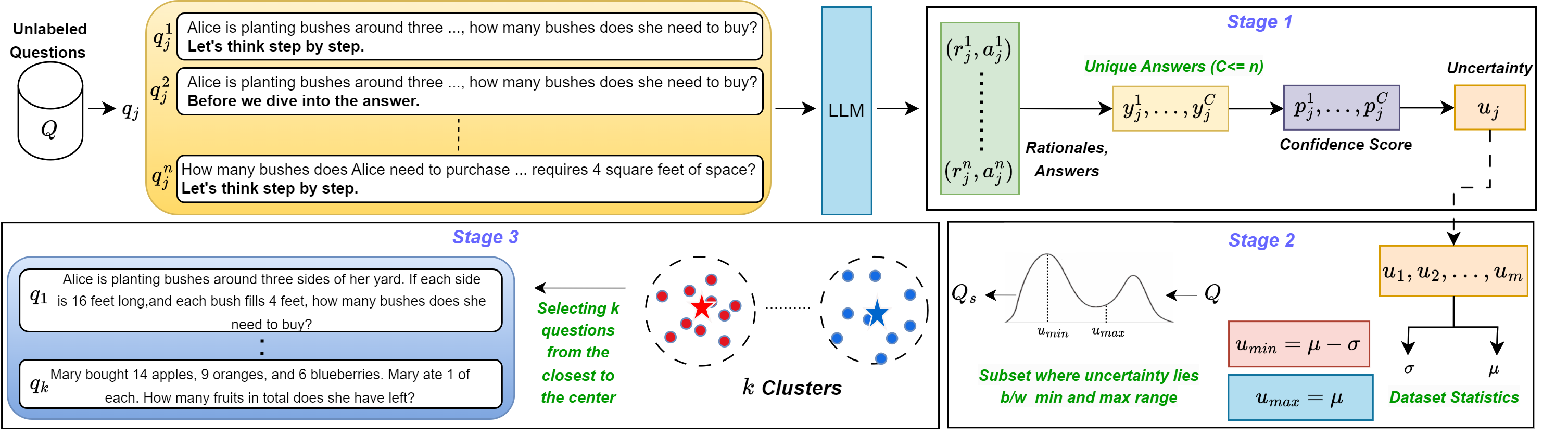}
    \caption{Overview of \textit{ZEUS}: Uncertainty for a question $q_j$ is calculated using a pool of answers generated using various prompts, including trigger phrases, non-zero temperature-based decoding, and rephrasing of $q_j$. Subsequently, questions with uncertainty within a certain range are selected and used for constructing demonstrations.}
    \label{fig:diagram}
\end{figure*}

\section{\textit{ZEUS}: \underline{Ze}ro-shot \underline{U}ncertainty-based \underline{S}election}
We propose the \textit{ZEUS} method, which aims to construct useful demonstrations containing a specific level of required uncertainty. It is comprised of three stages: (i) uncertainty estimation, (ii) uncertainty-based question selection, and (iii) demonstration construction. We have illustrated all the stages of ZEUS in Figure \ref{fig:diagram}.

\subsection{Uncertainty Estimation (Stage 1)}

In the \textit{ZEUS} method, uncertainty estimation is a critical step and is performed using perturbation. We exploit three distinct types of perturbations to estimate uncertainty for each unlabeled question in the set $Q$. These perturbations include temperature adjustments, trigger phrase variations, and question rephrasing.

\hspace{-1em}\textbf{Temperature Perturbation}: This perturbation technique is based on the principle that a question can be answered in multiple ways, and these variations can be explored by adjusting the temperature parameter during decoding. Temperature perturbation helps in simulating different reasoning paths within the LLM. When the temperature is set to a higher value, the model’s outputs become more diverse, while a lower temperature typically results in more confident and consistent responses \cite{koehn2009statistical}. According to \citet{wang2022self}, if an LLM is confident in its answer to a question $q_j$, the responses generated at various temperatures should reach the same answer. In contrast, if the LLM is uncertain, different temperatures will yield a range of potentially inconsistent answers. To estimate uncertainty using this property, we generate $n$ responses for a question $q_j$ by using the highest temperature (=1). These responses $\{{r^l_{j}}\}_{l=1}^n$ form the basis for our temperature-based uncertainty estimation.

\hspace{-1em}\textbf{Trigger Phrase Perturbation}: This factor leverages the sensitivity of LLM performance to trigger phrases. \citet{NEURIPS2022_8bb0d291} demonstrated that appending different trigger phrases to a question can affect the LLM’s output. By introducing variations in trigger phrases, we can assess whether the LLM’s responses remain consistent. If the LLM provides the same answer across different trigger phrases, it suggests a high level of confidence in its response. Conversely, varying answers across trigger phrases indicate that the question $q_j$ is challenging or that the LLM is uncertain. To apply this perturbation, we append a set of $t$ different trigger phrases to the original question $q_j$ and generate a corresponding set of responses $\{{r^l_{j}}\}_{l=1}^t$.

\hspace{-1em}\textbf{Rephrasing Perturbation}: The third technique utilizes rephrasing of the input question to explore the impact on the LLM’s responses. We hypothesize that if the LLM is confident about its answer, rephrasing the question should not significantly alter the generated answer. On the other hand, if the LLM's answer is influenced by specific biases or ambiguities in the original question, rephrasing may lead to a different response. To estimate uncertainty using rephrasing, we generate $v$ rephrased versions of the question $q_j$ and obtain the sets of responses $\{{r^l_{j}}\}_{l=1}^v$. 

By integrating these three types of perturbations—temperature adjustment, trigger phrase variation, and question rephrasing, we generate a diverse set of responses for each question $q_j$. Specifically, we produce a total of $n \times t \times v$ responses. This pool of answers reflects variations due to different decoding settings, trigger phrases, and question rephrasing, serving as Monte Carlo samples from the LLM’s likelihood distribution. From these responses, we identify $C$ $(\leq n)$ unique answers $y^1_{j}, \ldots, y^C_{j}$ for the question $q_j$. The confidence score $p(y^c_{j}|q_j)$ for each unique answer $y^c_{j}$ is computed based on the consistency of responses across the different perturbations. This score quantifies the degree of certainty associated with each answer and serves as a basis for selecting informative demonstrations in subsequent stages of the \textit{ZEUS} method. The confidence score $p(y^c_{j}|q_j)$ for a unique answer $y^c_{j}$ is defined as:

 \begin{equation} \label{eqn
} p(y^c_{j}|q_j) = \frac{1}{n}\sum_{l=1}^{n} 1(y^c_{j} = a^l_{j}) \end{equation}

where $1(\cdot)$ is the indicator function that evaluates to 1 if $y^c_{j}$ matches $a^l_{j}$ and 0 otherwise.

To represent the uncertainty of the LLM regarding the question $q_j$, we use predictive entropy (\textit{PE}) \cite{kumar2022diversity}. \textit{PE} is maximized when confidence scores are uniformly distributed across many unique answers and increases as the number of unique answers grows. It reaches zero when all answers are identical. The \textit{PE} for the question $q_j$ is computed as follows:

 \begin{equation} \label{eqn
} u_j = -\sum_{c=1}^{C} p(y^c_{j}|q_j) \cdot \log(p(y^c_{j}|q_j)) \end{equation}

where $u_j$ measures the degree of uncertainty by quantifying the diversity of the answers.

\begin{figure*}[!]
    \centering
    \includegraphics[scale=0.3]{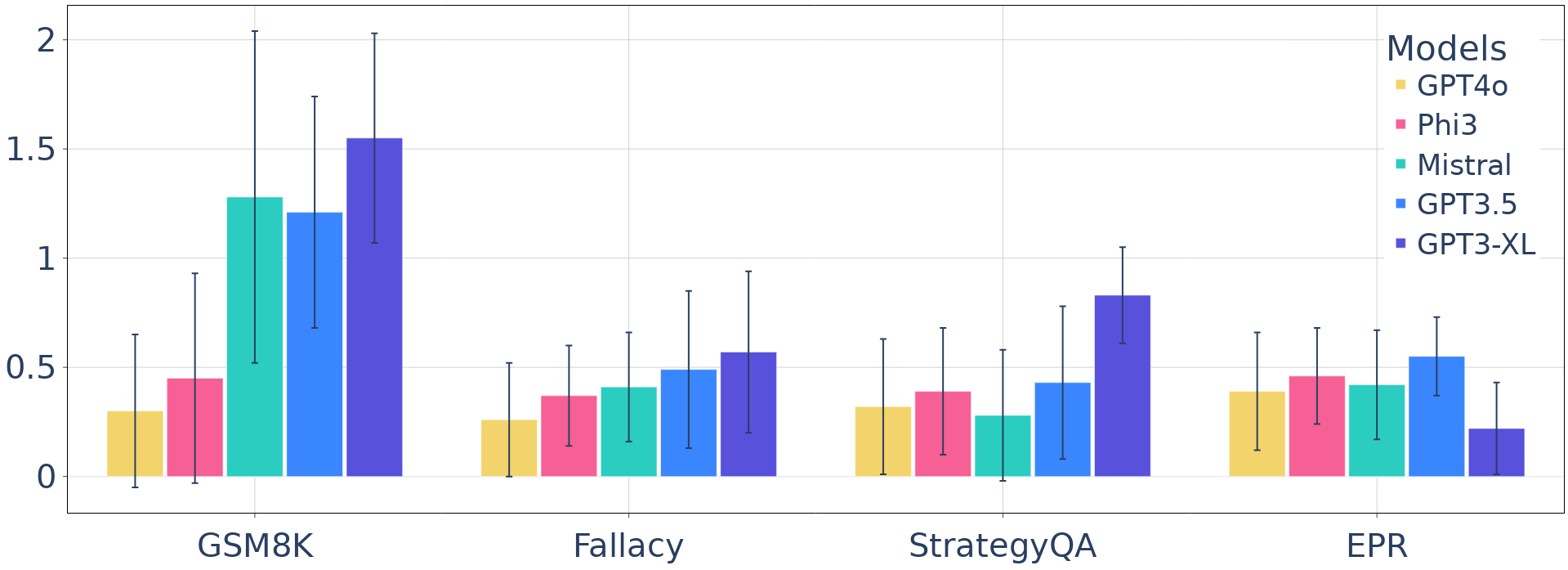}
    \textbf{}\caption{Mean and standard deviation of uncertainty values as error graph -specific statistics across models.}
    \label{fig:dataset_uncertainty}
\end{figure*}

\subsection{Uncertainty-based Selection (Stage 2)}
We define the LLM's overall understanding of the task using the mean uncertainty $\mu$ and the standard deviation $\sigma$ of the uncertainty estimates from the unlabeled set $Q$. A higher mean $\mu$ indicates a more challenging task for the LLM, while a higher standard deviation $\sigma$ reflects greater variability in question difficulty within $Q$. These two parameters provide insight into the usefulness of a question for improving the LLM's performance.

For instance, we hypothesize that when the mean uncertainty $\mu$ is low (indicating the LLM is performing well on the task), selecting questions with uncertainties lower than $\mu$ would not contribute valuable information. On the other hand, when the mean $\mu$ is high (suggesting the LLM struggles with the task), selecting questions with uncertainties significantly higher than $\mu$ may lead to less informative or erroneous rationales.

Based on these assumptions, we propose selecting a subset of questions $Q_s$ that fall within a specific uncertainty range, as defined by the following condition:

\begin{equation} \label{eqn}
\begin{aligned}
Q_s \subset Q = \{ q_j \mid u_{\min} \leq u_j < u_{\max} \}
\end{aligned}
\end{equation}

Here, $u_{\min}$ and $u_{\max}$ represent the minimum and maximum uncertainty thresholds used to select questions. In the subsequent section, we will detail the specific ranges (cf. Table~\ref{tbl:select_strategy}) based on $\mu$ and $\sigma$ for constructing demonstrations.

 \begin{table}[!]
\centering
\small
\setlength\tabcolsep{9pt}%
\begin{tabular}{*{200}{c}}
\toprule
{ \textbf{Strategy}} & \textbf{$u_{\min}$} & \textbf{$u_{\max}$}\\ 
  \midrule
{ \textit{Trivial}} &  0 & $\mu$ - $\sigma$  \\
{ \textit{Very Easy}} &  0 & $\mu$  \\
{ \textit{Easy}} &  0 & $\mu$ + $\sigma$\\
{ \textit{Moderate}} &  $\mu$ -  $\sigma$  & $\mu$ \\
{ \textit{Challenging}} &  $\mu$ - $\sigma$ & $\mu$ +$\sigma$  \\
{ \textit{Hard}} &  $\mu$ -  $\sigma$ & $\infty$ \\
{ \textit{Very Hard}} &  $\mu$ & $\infty$ \\
 \bottomrule 
\end{tabular}
\caption {Selection Strategies used in \textit{ZEUS} with their minimum $\mu_{\min}$ and maximum $\mu_{\max}$ range.}
  \label{tbl:select_strategy}
 \end{table}

\subsection{Demonstration Construction (Stage 3)}
We adopt the demonstration construction methodology from Auto-CoT, which emphasizes diversity to mitigate the influence of incorrect rationales generated by the Zero-Shot-CoT method. The selected questions $Q_s$ are first encoded into vector representations using Sentence Transformers \cite{reimers-gurevych-2019-sentence}. These vectors are then clustered using k-Means++ \cite{arthur2007k}, forming $k$ distinct clusters. From each cluster, the question closest to the cluster centroid is selected. The associated rationale and answer, generated by the Zero-Shot-CoT method, are then combined to form the demonstration set $D$. During inference, a test question $q$ is appended to the constructed demonstration $D$ and passed to the LLM for final predictions.

\section{Experimental Setup}

\textbf{Datasets}: We evaluate our proposed method on four challenging reasoning datasets. \textbf{GSM8K} \cite{cobbe2021training} comprises arithmetic reasoning problems. \textbf{StrategyQA} \cite{geva-etal-2021-aristotle} is a question-answering benchmark requiring implicit multi-hop reasoning. \textbf{Logical Fallacy} (referred to as \textit{Fallacy}) \cite{jin2022logical} involves reasoning about arguments and detecting formal and informal fallacies. \textbf{Epistemic Reasoning} (\textit{EPR}) \cite{sileo2023mindgames} is a natural language inference task that challenges LLMs to reason about human mental states. For a fair comparison, we split all datasets, except GSM8K, into two sets using stratified sampling: (i) an unlabeled set (70\%) for demonstration creation, and (ii) a test set (30\%) for zero-shot performance evaluation. GSM8K already contains train and test sets, so no further split was needed.

\hspace{-1em}\textbf{Implementation}: We conduct experiments using five LLMs: GPT-4o \cite{openai_gpt4o_2024}, \textit{Mistral-7B-Instruct-v0.2} (Mistral) \cite{jiang2023mistral}, \textit{Phi-3-mini-4k-instruct} (Phi3) \cite{abdin2024phi}, \textit{text-davinci-002} (GPT3-XL), and \textit{text-davinci-003} (GPT3.5) \cite{NEURIPS2020_1457c0d6}. Note that this models are including both open-source (Phi3, Mistral) and proprietary models (GPT-4o, GPT3.5, GPT3-XL). 
To ensure consistency with prior work such as Auto-CoT, we use $k=8$ demonstrations for all datasets, except for StrategyQA, where we use $k=6$. Additionally, during the evaluation of the LLMs, we set the temperature to 0 to ensure deterministic outputs, and report the average performance across three runs to maintain consistency in predictions.

\begin{figure}[!]
    \centering
    \includegraphics[scale=0.1]{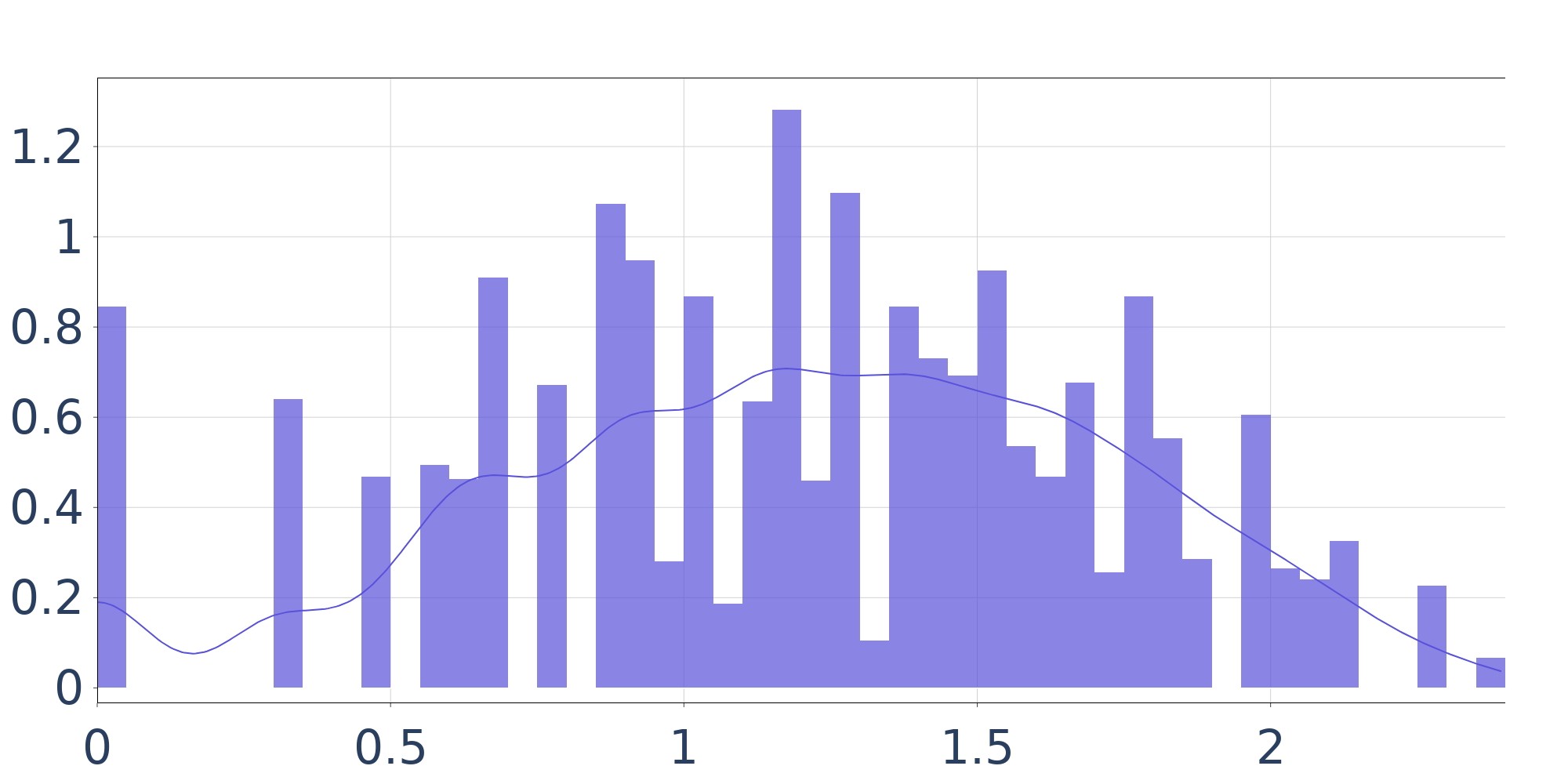}
    \textbf{}\caption{Probability density function of uncertainty estimates of our method using GPT3.5 on GSM8K.}
    \label{fig:gsm8k_zeroshot_entropy}
\end{figure}

\hspace{-1em}\textbf{Uncertainty Estimation in \textit{ZEUS}}: Uncertainty in \textit{ZEUS} is estimated using a combination of three perturbation methods: (1) non-zero temperature decoding, (2) trigger phrase variation, and (3) question rephrasing. We use five trigger phrases: " " (Empty), "\textit{Let's think step by step.}" (SS), "\textit{Let’s think about this logically step by step.}" (LSS), "\textit{Before we dive into the answer,}" (BDA), and "\textit{Before answering the question, let's understand the input.}" (BQU). For each question, we generate two rationale-answer pairs per trigger phrase at a temperature of 1, producing 10 rationale-answer pairs.

Each question is also rephrased using the instruction "\textit{Rephrase the below passage}" with GPT4o.\footnote{In general, rephrasing ensures that the intent of the question does not change.} We then generate five additional rationale-answer pairs using these rephrased questions with trigger phrases at a temperature of 0 to ensure precise responses. Thus, a total of 15 rationale-answer pairs are generated for each question to estimate uncertainty.

\hspace{-1em}\textbf{Selection Strategies in \textit{ZEUS}}: We define seven selection strategies based on the mean $\mu$ and standard deviation $\sigma$ of uncertainty values across the unlabeled set, detailed in Table \ref{tbl:select_strategy}. These strategies include: \textit{\textbf{Trivial}}, \textit{\textbf{Very Easy}}, and \textit{\textbf{Easy}} (selecting the lowest uncertainty demonstrations), \textit{\textbf{Challenging}}, \textit{\textbf{Hard}}, and \textit{\textbf{Very Hard}} (focusing on high uncertainty values), and \textbf{\textit{Moderate}} (selecting demonstrations from a range of uncertainty levels around $\mu$).

\hspace{-1em}\textbf{Baselines}: We compare \textit{ZEUS} against five baseline methods: Zero-Shot, Few-Shot,\footnote{Zero-Shot and Few-Shot baselines do not use rationales or trigger phrases, instead utilizing either zero or a few examples.} Zero-Shot-CoT \cite{NEURIPS2022_8bb0d291}, Manual-CoT (Few-Shot-CoT) \cite{wei2022chain}, and Auto-CoT.

\section{Results \& Qualitative  Analysis}

\subsection{Uncertainty Distribution Analysis}
In this subsection, we present an analysis of the mean ($\mu$) and standard deviation ($\sigma$) of uncertainty estimates for different LLMs across various reasoning datasets. In Figure \ref{fig:gsm8k_zeroshot_entropy}, we illustrate the distribution of uncertainty estimates for GPT-3.5 on the GSM8K dataset. We have provided the comprehensive plots of the distributions in the appendix (see Figures \ref{fig:uncertainty_gpt4o} --\ref{fig:uncertainty_gpt3xl}). The mean $\mu$ and standard deviation $\sigma$ of the uncertainty estimates using the unlabeled set $Q$ has been shown through an error bar graph in Figure \ref{fig:dataset_uncertainty}. Notably, LLMs such as GPT3-XL and Mistral show higher uncertainty in GSM8K, particularly with a larger deviation, whereas for tasks like StrategyQA and EPR, the uncertainty is generally more consistent across models, with GPT4o displaying the lowest variation. The trend highlights that model uncertainty is highly task-dependent, with complex reasoning tasks eliciting higher variability in predictions.

\begin{figure*}[!]
\begin{subfigure}{0.48\textwidth}
\centering   \captionsetup{justification=centering} 
  \includegraphics[scale=0.2]{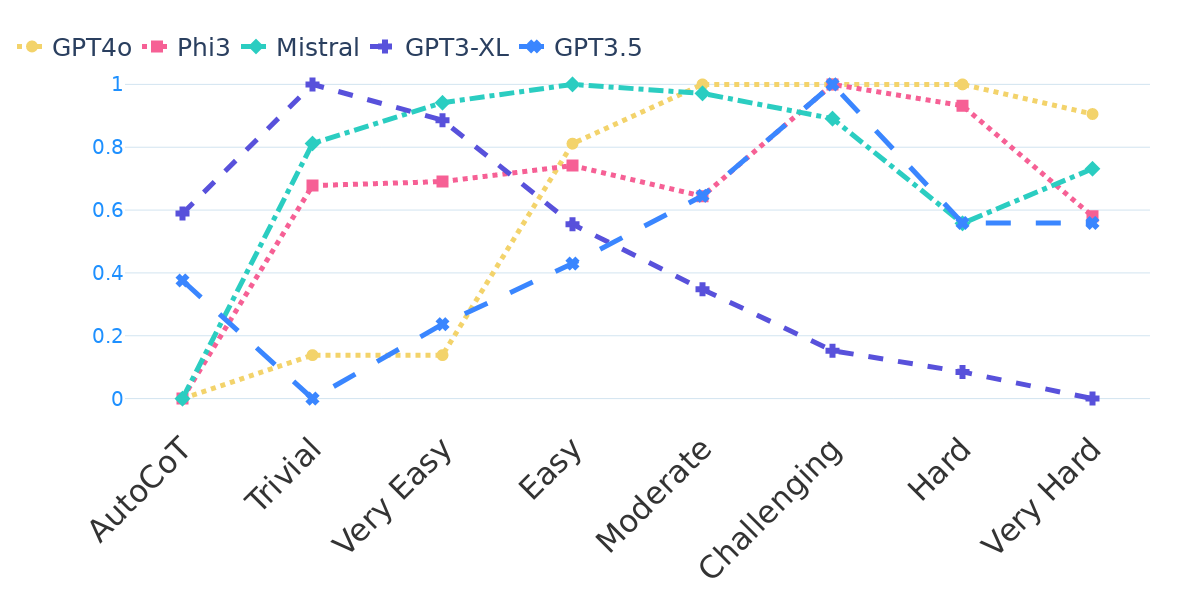}
  \caption{GSM8K}
\end{subfigure}
\hfill
\begin{subfigure}{0.48\textwidth}
  \centering   \captionsetup{justification=centering} 
  \includegraphics[scale=0.2]{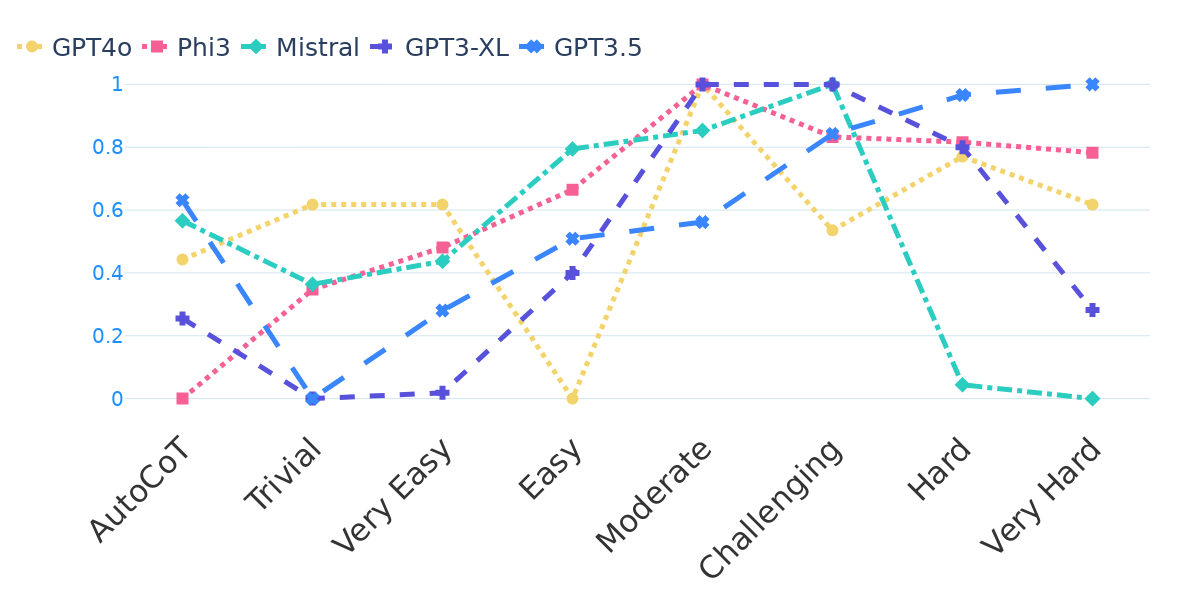}
  \caption{Fallacy}
\end{subfigure}
\hfill
\begin{subfigure}{0.48\textwidth}
  \centering   \captionsetup{justification=centering} 
  \includegraphics[scale=0.2]{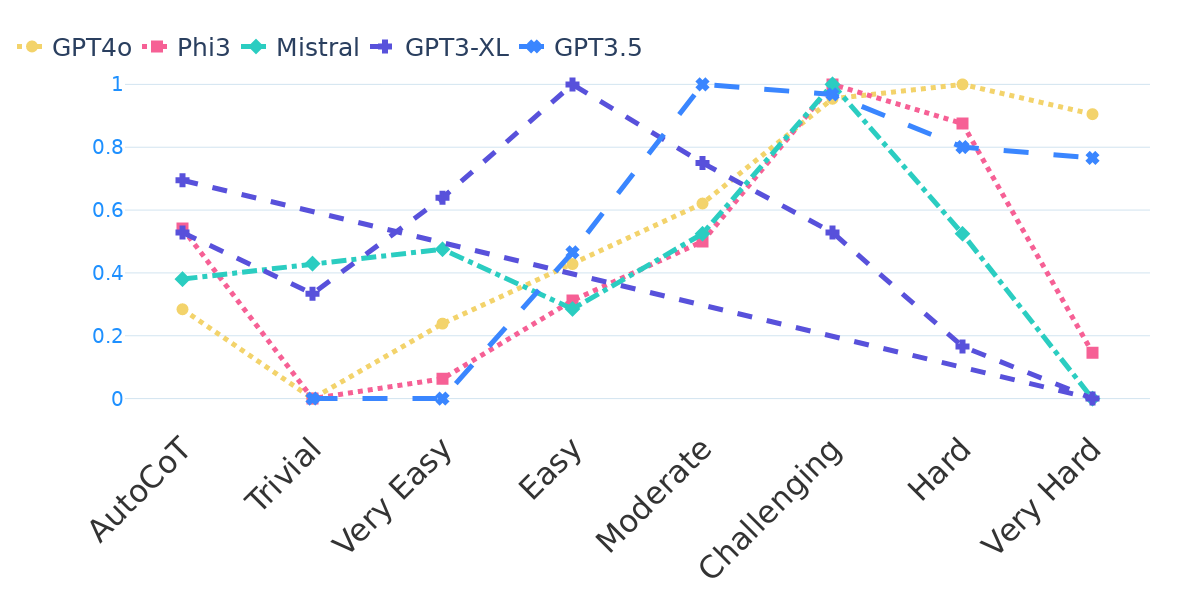}
  \caption{StrategyQA}
\end{subfigure}
\hfill
\begin{subfigure}{0.48\textwidth}
  \centering   \captionsetup{justification=centering} 
  \includegraphics[scale=0.2]{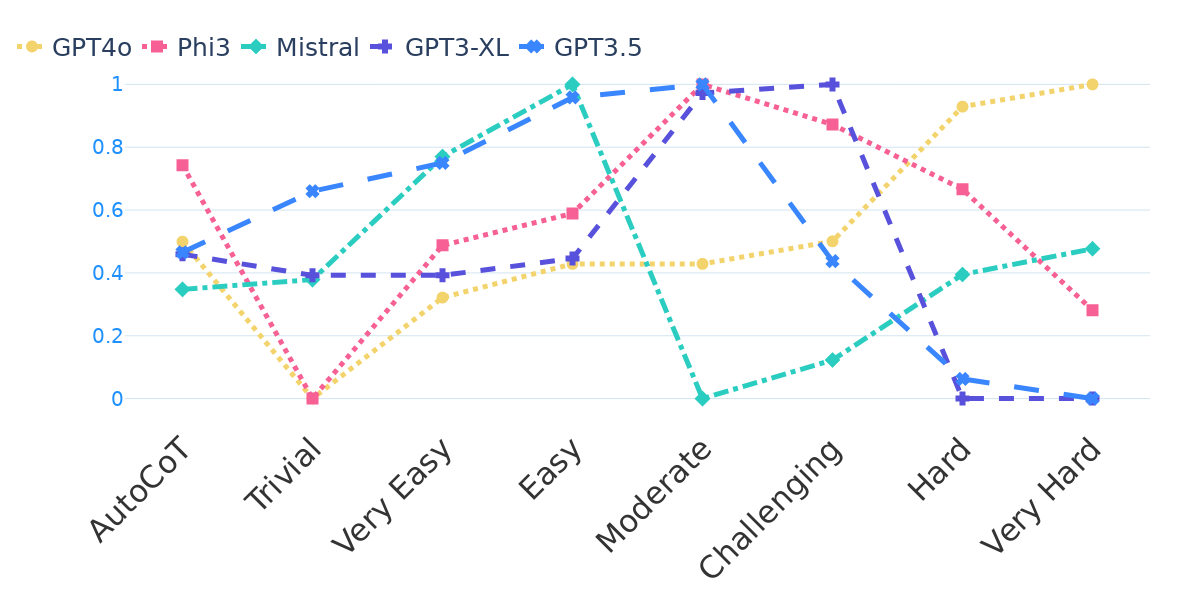}
  \caption{EPR}
\end{subfigure}
\caption{Normalized values of accuracy for various selection strategies using multiple LLMs.}
\label{fig:accuracy_dataset_plot}
\end{figure*}

\begin{figure}[!]
    \centering
    \includegraphics[scale=0.33]{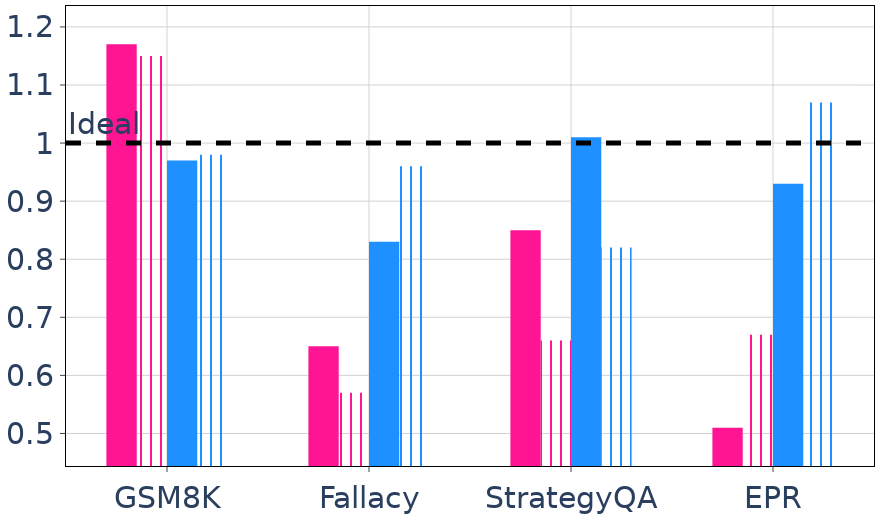}
    \caption{Sensitivity coefficient of confidence score wrt accuracy. {\color{blue} Blue} indicates \textit{ZEUS} and {\color{magenta}Magenta} for \textit{Temp-Perb}. Solid for GPT3-XL and Dashed for GPT3.5. Coefficient using \textit{ZEUS} is closest to ideal coefficient.}
    \label{fig:correlation}
\end{figure}

\subsection{Sensitivity of Uncertainty Estimates}
To assess the sensitivity of uncertainty estimates in distinguishing between helpful and redundant questions, we investigate the relationship between confidence scores and accuracy. This is done by fitting a linear regression (LR) model between the confidence score of the most common answer and its corresponding accuracy. In an ideally sensitive model, the slope coefficient of the LR would be one, indicating that a unit change in confidence directly corresponds to a unit change in accuracy. We compare our confidence scoring method against a temperature-based perturbation approach ~\cite{wan-etal-2023-better, diao2023active, gao2024spuq}, referred to as \textit{Temp-Perb}. This comparison is carried out using Zero-Shot-CoT prompting with 15 distinct temperature perturbations.

Figure \ref{fig:correlation} shows the slope coefficients for both \textit{ZEUS} and \textit{Temp-Perb}. Our results demonstrate that \textit{ZEUS} consistently produces slope coefficients closer to the ideal sensitivity compared to \textit{Temp-Perb}. Interestingly, \textit{Temp-Perb} shows notably low sensitivity in the Logical Fallacy and EPR datasets, indicating a lack of reliability. In contrast, for GSM8K, \textit{Temp-Perb} exhibits a coefficient exceeding 1, reflecting excessive sensitivity in this task.

\begin{table*}[ht]
\centering
\small
\setlength\tabcolsep{5pt}
\begin{tabular}{c|cc|cc|cc|cc}
\toprule
\multicolumn{1}{c}{} & \multicolumn{2}{c}{GSM8K}  & \multicolumn{2}{|c}{Fallacy}   & \multicolumn{2}{|c}{StrategyQA}   & \multicolumn{2}{|c}{EPR}\\
\cmidrule(lr){2-3}\cmidrule(lr){4-5}\cmidrule(lr){6-7}\cmidrule(lr){8-9}
 \multirow{-1}{*}{\textbf{Model}} & \textbf{Best}  & \textbf{Worst} & \textbf{Best}  & \textbf{Worst} & \textbf{Best}  & \textbf{Worst} & \textbf{Best}  & \textbf{Worst}  \\ 
\midrule
GPT4o      & \textcolor{blue!60!green}{Hard}       & \textcolor{red!70!black}{Trivial}     & \textcolor{blue!60!green}{Moderate}     & \textcolor{red!70!black}{Easy}          & \textcolor{blue!60!green}{Hard}             & \textcolor{red!70!black}{Trivial}           & \textcolor{blue!60!green}{Very Hard} & \textcolor{red!70!black}{Trivial} \\
Phi3       & \textcolor{blue!60!green}{Challenging} & \textcolor{red!70!black}{Very Hard}   & \textcolor{blue!60!green}{Moderate}     & \textcolor{red!70!black}{Trivial}        & \textcolor{blue!60!green}{Challenging}       & \textcolor{red!70!black}{Trivial}           & \textcolor{blue!60!green}{Moderate}  & \textcolor{red!70!black}{Trivial} \\
Mistral    & \textcolor{blue!60!green}{Easy}       & \textcolor{red!70!black}{Hard}        & \textcolor{blue!60!green}{Challenging}  & \textcolor{red!70!black}{Very Hard}      & \textcolor{blue!60!green}{Challenging}          & \textcolor{red!70!black}{Very Hard}           & \textcolor{blue!60!green}{Easy}      & \textcolor{red!70!black}{Moderate} \\
GPT3.5     & \textcolor{blue!60!green}{Challenging} & \textcolor{red!70!black}{Trivial}     & \textcolor{blue!60!green}{Very Hard}    & \textcolor{red!70!black}{Trivial}        & \textcolor{blue!60!green}{Moderate}              & \textcolor{red!70!black}{Trivial}           & \textcolor{blue!60!green}{Moderate}  & \textcolor{red!70!black}{Very Hard} \\
GPT3-XL    & \textcolor{blue!60!green}{Trivial}     & \textcolor{red!70!black}{Very Hard}   & \textcolor{blue!60!green}{Challenging}  & \textcolor{red!70!black}{Trivial}    & \textcolor{blue!60!green}{Easy}           & \textcolor{red!70!black}{Very Hard}           & \textcolor{blue!60!green}{Challenging} & \textcolor{red!70!black}{Very Hard} \\
\bottomrule
\end{tabular}
\caption{Best and worst-performing strategies across tasks for each model, indicating that GPT4o requires harder strategies for optimal performance, while GPT3-XL shows improved results with easier strategies.}
\label{tbl:trend}
\end{table*}

\begin{figure*}[!]
\begin{subfigure}{0.48\textwidth}
\centering   \captionsetup{justification=centering} 
  \includegraphics[scale=0.2]{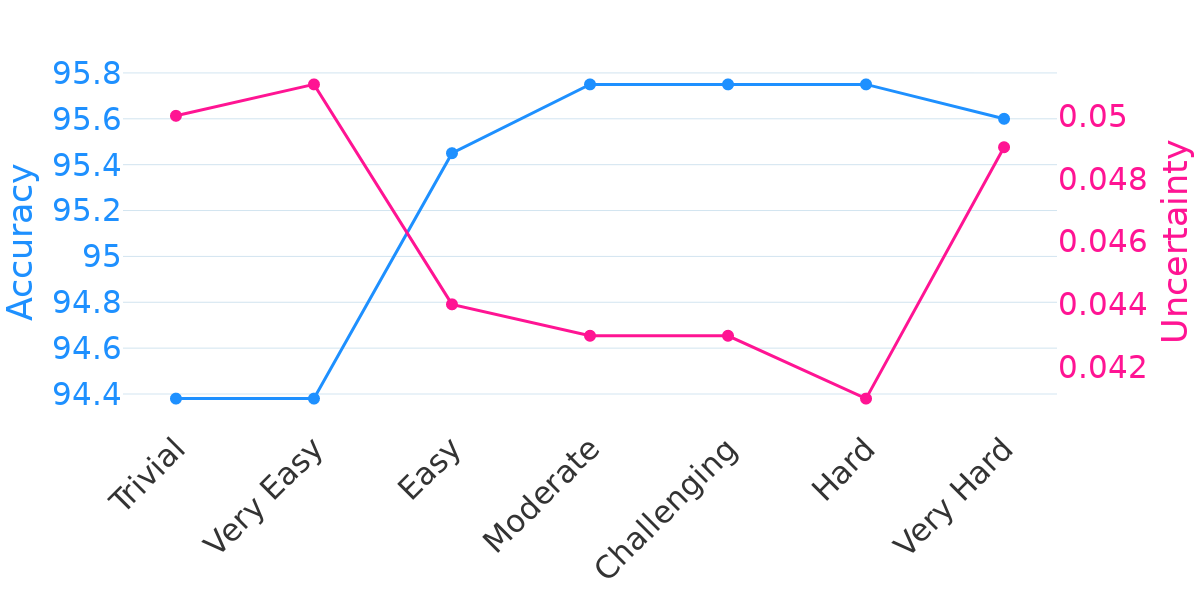}
  \caption{GSM8K}
\end{subfigure}
\hfill
\begin{subfigure}{0.49\textwidth}
  \centering   \captionsetup{justification=centering} 
  \includegraphics[scale=0.2]{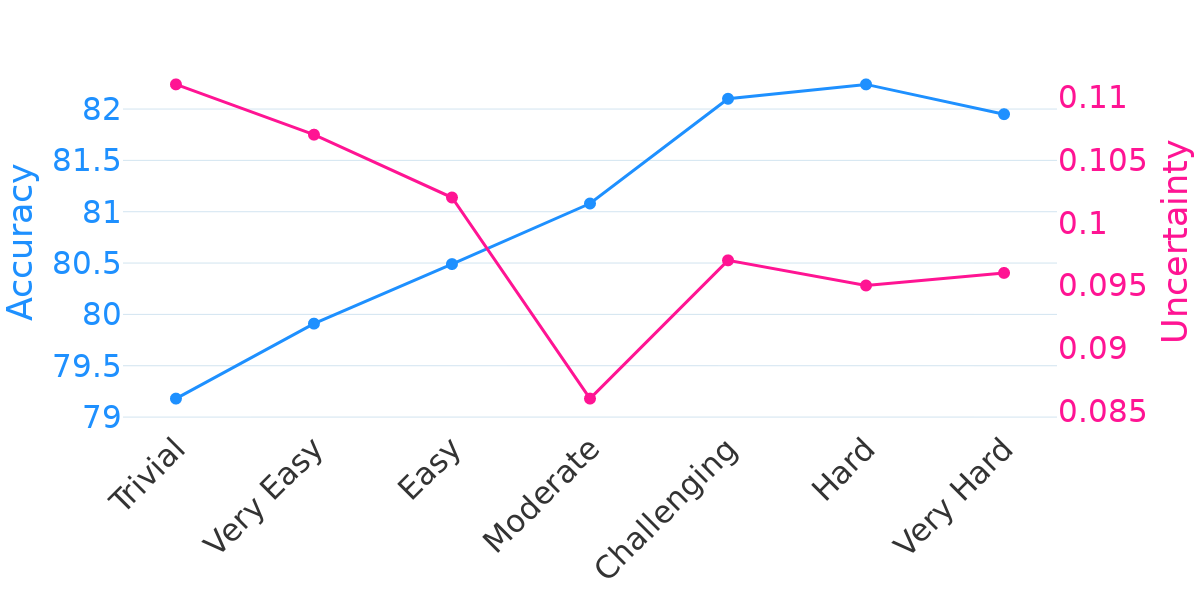}
  \caption{StrategyQA}
\end{subfigure}
\caption{Accuracy vs \textit{Temp-Perb} Uncertainty trend across all selection strategies for \textbf{GPT4o}.}\label{fig:accuracy_vs_uncertainty}
\end{figure*}

\subsection{Analysis of Selection Strategies}
We present the normalized accuracy values for all selection strategies, including AutoCoT, in Figure \ref{fig:accuracy_dataset_plot}. Our analysis reveals that AutoCoT was consistently outperformed by at least one other strategy across all LLMs and datasets. This indicates that leveraging uncertainty-based demonstration creation can more effectively identify valuable questions that enhance model performance. To provide a clearer perspective, Table \ref{tbl:trend} details the best and worst selection strategies for each model and dataset.

LLMs exhibit distinct performance patterns across varying levels of question difficulty, which allows us to categorize them into two broad groups: advanced models (GPT-4o, Phi3, GPT-3.5) and simpler models (Mistral, GPT-3 XL). This classification is based on observed performance trends rather than model size or architecture alone. Advanced models excel in handling \textit{Hard} and \textit{Very Hard} questions due to their superior reasoning capabilities, but they show limited gains when engaging with \textit{Trivial} or \textit{Very Easy} strategies, where their advanced abilities remain underutilized. On the other hand, simpler models perform better with \textit{Trivial} and \textit{Easy} strategies, as these align well with their baseline capabilities. However, they struggle considerably with \textit{Hard} and \textit{Very Hard} questions, where errors and uninformative outputs become more prevalent.

To capture general trends, we analyzed performance across the best and worst strategies for each model. Our findings highlight that while \textit{Trivial} and \textit{Very Easy} strategies consistently yield the lowest performance for advanced models, simpler models face significant challenges with \textit{Hard} and \textit{Very Hard} strategies. Notably, our categorization focuses on overall performance trends rather than model size, which places models like GPT-4o and Phi3 in the same group. 

Among the selection strategies, \textit{Trivial} and \textit{Very Hard} tend to yield poorer performance across most models. This suggests that extremes in task difficulty—whether too easy or too hard—are generally detrimental to model accuracy. The \textit{Hard} strategy generally improves performance for GPT-4o, whereas the \textit{Challenging} strategy appears to be optimal for Phi3, Mistral, and GPT-3.5. These findings align with the overall performance trends observed for these models.

However, performance variations still exist across tasks and models. For instance, the Mistral model's performance declines with \textit{Moderate} and harder strategies on the EPR task, while it improves with higher uncertainty estimates on other tasks. This indicates that selecting the optimal strategy can be complex and task-dependent. To address this, the next subsection will explore methods for determining the most effective selection strategy.

\begin{table*}[ht]
\centering
\small
\setlength\tabcolsep{4pt}
\begin{tabular}{c|ccccc|ccccc}
\toprule
\multicolumn{1}{c}{} & \multicolumn{5}{c}{GSM8K}  & \multicolumn{5}{|c}{Fallacy}\\
    \cmidrule(lr){2-6}\cmidrule(lr){7-11}
 \multirow{-1}{*}{\textbf{Method}} & GPT4o  & Phi3 & Mistral  & GPT3.5 & GPT3-XL & GPT4o  & Phi3 & Mistral  & GPT3.5 & GPT3-XL\\ 
\midrule
Zero-Shot & 49.4 & 50.7 & 45.3 & 12.6 & 10.7 & 80.5 & 81.8 & 71.1 & 63.9  & 48.3\\
Few-Shot & 84.0 & 50.7 & 45.3 & 16.5 & 14.4 & 92.5 & 90.4 & 62.9 & 76.9 & \textbf{79.8} \\
Zero-Shot-CoT & 94.8 & 85.9 & 51.8 & 60.4 & 44.7 & 84.8 & 87.5 & 67.1 & 67.7 & 61.7\\
Manual-CoT & 89.3 & 81.9 & 42.4 & 56.4 & 43.9 & 90.1 & 90.1 & 64.3 & - & -\\
Auto-CoT & 94.2 & 87.6 & 47.2 & 58.5 & 44.6 &  97.0 & 85.6 & 74.4 & 76.9 & 66.7
\\
\midrule
\textit{ZEUS} \textit{(LU)} & \textbf{95.8}  &  \textbf{89.9} & 57.3 & \textbf{62.9} & \textbf{51.9} & 98.0 & 94.0  & \textbf{78.5}  & \textbf{79.4}  & 76.4  \\
\textit{ZEUS} \textit{(HA)} & \textbf{95.8}  &  \textbf{89.9} & \textbf{57.6} & \textbf{62.9} & \textbf{51.9} & \textbf{98.0} & \textbf{94.0}  & \textbf{78.5}  & \textbf{79.4}  & 76.4  \\
\bottomrule
\toprule
\multicolumn{1}{c}{} & \multicolumn{5}{c}{StrategyQA}  & \multicolumn{5}{|c}{EPR}\\
    \cmidrule(lr){2-6}\cmidrule(lr){7-11}
 \multirow{-1}{*}{\textbf{}} & GPT4o  & Phi3 & Mistral  & GPT3.5 & GPT3-XL & GPT4o  & Phi3 & Mistral  & GPT3.5 & GPT3-XL\\ 
\midrule
Zero-Shot & 65.2 & 56.6 & 59.8 & 54.4 & 16.6 & 61.2 & 72.2 & 45.2 & 60.0  & 61.5 \\
Few-Shot & 77.6 & 65.8 & 61.1 & 66.2  &  64.8 & 83.0 & 64.0 & 55.2 & \textbf{75.6} & 58.2  \\
Zero-Shot-CoT & 70.7 & 67.5 & 59.0 & 57.4  &  51.2 &  64.7 & \textbf{79.8} & 65.7 & 60.2   &  59.3 \\
Manual-CoT & 81.1 & \textbf{68.9} & \textbf{63.8} & \textbf{68.6} & 57.6 & \textbf{84.2} & 64.0 & 57.7 & - & -\\
Auto-CoT & 80.1 & 64.5 & 57.9 & 64.9 & 64.1  & 68.2 &  75.3 & 52.5 & 52.5 &  59.5
\\
\midrule
\textit{ZEUS} \textit{(LU)} & 81.1  & 67.7 & 59.8 & 66.8 &  \textbf{66.5}  & 72.8 & 76.2 & \textbf{68.5}  & 65.3  & \textbf{66.2} \\
\textit{ZEUS} \textit{(HA)} & \textbf{82.2} & 67.7 & 59.8 & 66.8 &  \textbf{66.5}  & 72.8 & 77.0 & \textbf{68.5} & 65.3  &\textbf{66.2} \\
\bottomrule
\end{tabular}
\caption {Accuracy on various datasets. \textit{ZEUS} \textit{(HA)} chooses the best performing strategy for each dataset while \textit{ZEUS} \textit{(LU)} chooses the strategies having lowest \textit{Temp-Perb} uncertainty estimates.} 
\label{tbl:result}
\end{table*}

\subsection{Choosing Optimal Selection Strategy}
Upon constructing the demonstration for each strategy, we need to identify the optimal strategy for a given task and model. We calculate the average uncertainty on the unlabelled set Q while keeping the demonstration unchanged. The optimal strategy is the one with the lowest entropy, as this tends to strongly correlate with higher accuracy. \textit{Temp-Perb} provides well-calibrated uncertainty estimates, although it lacks the sensitivity required to effectively differentiate between similar questions. Despite this limitation, its well calibration makes \textit{Temp-Perb} suitable for selecting the best-performing strategy based on uncertainty estimates. Therefore, we use \textit{Temp-Perb} for uncertainty estimation to determine the optimal selection strategy for a given model and task.


In Figure \ref{fig:accuracy_vs_uncertainty}, we illustrate the accuracy of various selection strategies for GPT-4o in relation to \textit{Temp-Perb} based uncertainty estimates. The data indicates that the accuracy is inversely correlated with uncertainty across all four datasets. This inverse relationship allows us to identify the optimal selection strategy as the one associated with the lowest uncertainty. We have included similar analyses for other models in the appendix (cf. Figures \ref{fig:gpt4o_acc_entropy} -- \ref{fig:gpt3xl_acc_entropy}).

\subsection{Comparison with Baselines}
The selection strategy with the lowest uncertainty is denoted as \textit{ZEUS (LU)}, while the strategy with the highest accuracy is represented by \textit{ZEUS (HA)}. Table \ref{tbl:result} demonstrates that \textit{ZEUS (LU)} and \textit{ZEUS (HA)} yield nearly identical performance, underscoring the robustness of the \textit{Temp-Perb} uncertainty estimates. In general, the optimally selected \textit{ZEUS(LU)} either outperforms all baseline methods or comes in a close second to in a few cases across three datasets (GSM8K, Fallacy, and Strategy QA), with only a few exceptions. \textit{ZEUS} methods consistently outperform all baseline strategies on the GSM8K and Fallacy datasets, with the exception of GPT-3 XL on the Fallacy dataset. For the StrategyQA dataset, Manual-CoT achieves the highest accuracy for most models, highlighting the effectiveness of human-crafted demonstrations. On the EPR dataset, \textit{ZEUS} surpasses Zero-Shot, Zero-Shot-CoT, and Auto-CoT methods across most models. Overall, \textit{ZEUS} methods either match or exceed the accuracy of these baseline strategies without requiring manual annotations.

\section{Conclusion} 
This paper introduces the zero-shot uncertainty-based \textit{ZEUS} method for evaluating and selecting optimal strategies based on uncertainty estimates. Our analysis reveals that \textit{ZEUS} provides highly sensitive and reliable uncertainty estimates, outperforming temperature-based perturbation approaches (\textit{Temp-Perb}) in distinguishing between helpful and redundant questions.

Our findings classify models into two groups based on their optimal strategies. Advanced models like GPT-4o, Phi3, and GPT3.5 perform best with \textit{Hard} and \textit{Challenging} example selection strategies, effectively leveraging their greater capabilities to tackle complex queries. In contrast, simpler models such as Mistral and GPT3-XL benefit more from \textit{Trivial} and \textit{Easy} strategies, where even low-uncertainty questions yield valuable information. By selecting the strategy with the lowest uncertainty estimates, \textit{ZEUS(LU)} (recommended) achieves performance comparable to the best-performing strategies \textit{ZEUS(HA)}, without requiring manual annotations. Overall, \textit{ZEUS} consistently matches or surpasses baseline accuracy, demonstrating its robustness and sensitivity in improving model performance.

\section{Limitation}
While our work demonstrates the effectiveness of the \textit{ZEUS} method, there are several limitations and avenues for future research. First, the selection strategies in our current approach require exhaustive exploration to find the optimal strategy, which can be time-consuming and computationally expensive. This process could be automated by incorporating a greedy search algorithm based on uncertainty estimates, allowing for more efficient strategy selection. Another limitation is our reliance on uncertainty estimates from unlabeled questions, without examining the impact of dataset attributes like diversity or size. These factors could affect the estimates and lead to suboptimal strategy selection. Future work should explore these effects to improve robustness.

\bibliography{custom}

\begin{thebibliography}{46}
\providecommand{\natexlab}[1]{#1}

\bibitem[{Abdin et~al.(2024)Abdin, Jacobs, Awan, Aneja, Awadallah, Awadalla, Bach, Bahree, Bakhtiari, Behl et~al.}]{abdin2024phi}
Marah Abdin, Sam~Ade Jacobs, Ammar~Ahmad Awan, Jyoti Aneja, Ahmed Awadallah, Hany Awadalla, Nguyen Bach, Amit Bahree, Arash Bakhtiari, Harkirat Behl, et~al. 2024.
\newblock Phi-3 technical report: A highly capable language model locally on your phone.
\newblock \emph{arXiv preprint arXiv:2404.14219}.

\bibitem[{Arthur and Vassilvitskii(2007)}]{arthur2007k}
David Arthur and Sergei Vassilvitskii. 2007.
\newblock K-means++ the advantages of careful seeding.
\newblock In \emph{Proceedings of the eighteenth annual ACM-SIAM symposium on Discrete algorithms}, pages 1027--1035.

\bibitem[{Bayer and Reuter(2024)}]{bayer2024activellm}
Markus Bayer and Christian Reuter. 2024.
\newblock Activellm: Large language model-based active learning for textual few-shot scenarios.
\newblock \emph{arXiv preprint arXiv:2405.10808}.

\bibitem[{Brown et~al.(2020)Brown, Mann, Ryder, Subbiah, Kaplan, Dhariwal, Neelakantan, Shyam, Sastry, Askell, Agarwal, Herbert-Voss, Krueger, Henighan, Child, Ramesh, Ziegler, Wu, Winter, Hesse, Chen, Sigler, Litwin, Gray, Chess, Clark, Berner, McCandlish, Radford, Sutskever, and Amodei}]{NEURIPS2020_1457c0d6}
Tom Brown, Benjamin Mann, Nick Ryder, Melanie Subbiah, Jared~D Kaplan, Prafulla Dhariwal, Arvind Neelakantan, Pranav Shyam, Girish Sastry, Amanda Askell, Sandhini Agarwal, Ariel Herbert-Voss, Gretchen Krueger, Tom Henighan, Rewon Child, Aditya Ramesh, Daniel Ziegler, Jeffrey Wu, Clemens Winter, Chris Hesse, Mark Chen, Eric Sigler, Mateusz Litwin, Scott Gray, Benjamin Chess, Jack Clark, Christopher Berner, Sam McCandlish, Alec Radford, Ilya Sutskever, and Dario Amodei. 2020.
\newblock \href {https://proceedings.neurips.cc/paper_files/paper/2020/file/1457c0d6bfcb4967418bfb8ac142f64a-Paper.pdf} {Language models are few-shot learners}.
\newblock In \emph{Advances in Neural Information Processing Systems}, volume~33, pages 1877--1901. Curran Associates, Inc.

\bibitem[{Cobbe et~al.(2021)Cobbe, Kosaraju, Bavarian, Chen, Jun, Kaiser, Plappert, Tworek, Hilton, Nakano et~al.}]{cobbe2021training}
Karl Cobbe, Vineet Kosaraju, Mohammad Bavarian, Mark Chen, Heewoo Jun, Lukasz Kaiser, Matthias Plappert, Jerry Tworek, Jacob Hilton, Reiichiro Nakano, et~al. 2021.
\newblock Training verifiers to solve math word problems.
\newblock \emph{arXiv preprint arXiv:2110.14168}.

\bibitem[{Diao et~al.(2024)Diao, Wang, Lin, Pan, Liu, and Zhang}]{diao-etal-2024-active}
Shizhe Diao, Pengcheng Wang, Yong Lin, Rui Pan, Xiang Liu, and Tong Zhang. 2024.
\newblock \href {https://aclanthology.org/2024.acl-long.73} {Active prompting with chain-of-thought for large language models}.
\newblock In \emph{Proceedings of the 62nd Annual Meeting of the Association for Computational Linguistics (Volume 1: Long Papers)}, pages 1330--1350, Bangkok, Thailand. Association for Computational Linguistics.

\bibitem[{Diao et~al.(2023)Diao, Wang, Lin, and Zhang}]{diao2023active}
Shizhe Diao, Pengcheng Wang, Yong Lin, and Tong Zhang. 2023.
\newblock Active prompting with chain-of-thought for large language models.
\newblock \emph{arXiv preprint arXiv:2302.12246}.

\bibitem[{Feng et~al.(2024)Feng, Zhang, Gu, Ye, He, and Wang}]{feng2024towards}
Guhao Feng, Bohang Zhang, Yuntian Gu, Haotian Ye, Di~He, and Liwei Wang. 2024.
\newblock Towards revealing the mystery behind chain of thought: a theoretical perspective.
\newblock \emph{Advances in Neural Information Processing Systems}, 36.

\bibitem[{Fu et~al.(2013)Fu, Zhu, and Li}]{fu2013survey}
Yifan Fu, Xingquan Zhu, and Bin Li. 2013.
\newblock A survey on instance selection for active learning.
\newblock \emph{Knowledge and information systems}, 35(2):249--283.

\bibitem[{Gal and Ghahramani(2016)}]{gal2016dropout}
Yarin Gal and Zoubin Ghahramani. 2016.
\newblock Dropout as a bayesian approximation: Representing model uncertainty in deep learning.
\newblock In \emph{Tnternational Conference on Machine Learning}, pages 1050--1059.

\bibitem[{Gao et~al.(2024)Gao, Zhang, Mouatadid, and Das}]{gao2024spuq}
Xiang Gao, Jiaxin Zhang, Lalla Mouatadid, and Kamalika Das. 2024.
\newblock Spuq: Perturbation-based uncertainty quantification for large language models.
\newblock In \emph{Proceedings of the 18th Conference of the European Chapter of the Association for Computational Linguistics (Volume 1: Long Papers)}, pages 2336--2346.

\bibitem[{Geva et~al.(2021)Geva, Khashabi, Segal, Khot, Roth, and Berant}]{geva-etal-2021-aristotle}
Mor Geva, Daniel Khashabi, Elad Segal, Tushar Khot, Dan Roth, and Jonathan Berant. 2021.
\newblock \href {https://doi.org/10.1162/tacl_a_00370} {Did aristotle use a laptop? a question answering benchmark with implicit reasoning strategies}.
\newblock \emph{Transactions of the Association for Computational Linguistics}, 9:346--361.

\bibitem[{Guo et~al.(2017)Guo, Pleiss, Sun, and Weinberger}]{pmlr-v70-guo17a}
Chuan Guo, Geoff Pleiss, Yu~Sun, and Kilian~Q. Weinberger. 2017.
\newblock \href {https://proceedings.mlr.press/v70/guo17a.html} {On calibration of modern neural networks}.
\newblock In \emph{Proceedings of the 34th International Conference on Machine Learning}, volume~70 of \emph{Proceedings of Machine Learning Research}, pages 1321--1330. PMLR.

\bibitem[{Hendrycks and Gimpel(2016)}]{hendrycks2016baseline}
Dan Hendrycks and Kevin Gimpel. 2016.
\newblock A baseline for detecting misclassified and out-of-distribution examples in neural networks.
\newblock In \emph{International Conference on Learning Representations}.

\bibitem[{Jiang et~al.(2023)Jiang, Sablayrolles, Mensch, Bamford, Chaplot, Casas, Bressand, Lengyel, Lample, Saulnier et~al.}]{jiang2023mistral}
Albert~Q Jiang, Alexandre Sablayrolles, Arthur Mensch, Chris Bamford, Devendra~Singh Chaplot, Diego de~las Casas, Florian Bressand, Gianna Lengyel, Guillaume Lample, Lucile Saulnier, et~al. 2023.
\newblock Mistral 7b.
\newblock \emph{arXiv preprint arXiv:2310.06825}.

\bibitem[{Jin et~al.(2022)Jin, Lalwani, Vaidhya, Shen, Ding, Lyu, Sachan, Mihalcea, and Schoelkopf}]{jin2022logical}
Zhijing Jin, Abhinav Lalwani, Tejas Vaidhya, Xiaoyu Shen, Yiwen Ding, Zhiheng Lyu, Mrinmaya Sachan, Rada Mihalcea, and Bernhard Schoelkopf. 2022.
\newblock Logical fallacy detection.
\newblock In \emph{Findings of the Association for Computational Linguistics: EMNLP 2022}, pages 7180--7198.

\bibitem[{Koehn(2009)}]{koehn2009statistical}
Philipp Koehn. 2009.
\newblock \emph{Statistical machine translation}.
\newblock Cambridge University Press.

\bibitem[{Kojima et~al.(2022)Kojima, Gu, Reid, Matsuo, and Iwasawa}]{NEURIPS2022_8bb0d291}
Takeshi Kojima, Shixiang~(Shane) Gu, Machel Reid, Yutaka Matsuo, and Yusuke Iwasawa. 2022.
\newblock \href {https://proceedings.neurips.cc/paper_files/paper/2022/file/8bb0d291acd4acf06ef112099c16f326-Paper-Conference.pdf} {Large language models are zero-shot reasoners}.
\newblock In \emph{Advances in Neural Information Processing Systems}, volume~35, pages 22199--22213. Curran Associates, Inc.

\bibitem[{Kong et~al.(2024)Kong, Zhao, Chen, Li, Qin, Sun, Zhou, Wang, and Dong}]{kong-etal-2024-better}
Aobo Kong, Shiwan Zhao, Hao Chen, Qicheng Li, Yong Qin, Ruiqi Sun, Xin Zhou, Enzhi Wang, and Xiaohang Dong. 2024.
\newblock \href {https://doi.org/10.18653/v1/2024.naacl-long.228} {Better zero-shot reasoning with role-play prompting}.
\newblock In \emph{Proceedings of the 2024 Conference of the North American Chapter of the Association for Computational Linguistics: Human Language Technologies (Volume 1: Long Papers)}, pages 4099--4113, Mexico City, Mexico. Association for Computational Linguistics.

\bibitem[{Kuhn et~al.(2023)Kuhn, Gal, and Farquhar}]{kuhn2023semantic}
Lorenz Kuhn, Yarin Gal, and Sebastian Farquhar. 2023.
\newblock Semantic uncertainty: Linguistic invariances for uncertainty estimation in natural language generation.
\newblock \emph{arXiv preprint arXiv:2302.09664}.

\bibitem[{Kumar et~al.(2022)Kumar, Dandapat, and Choudhury}]{kumar2022diversity}
Shanu Kumar, Sandipan Dandapat, and Monojit Choudhury. 2022.
\newblock ” diversity and uncertainty in moderation” are the key to data selection for multilingual few-shot transfer.
\newblock In \emph{Findings of the Association for Computational Linguistics: NAACL 2022}, pages 1042--1055.

\bibitem[{Lakshminarayanan et~al.(2017)Lakshminarayanan, Pritzel, and Blundell}]{NIPS2017_9ef2ed4b}
Balaji Lakshminarayanan, Alexander Pritzel, and Charles Blundell. 2017.
\newblock \href {https://proceedings.neurips.cc/paper_files/paper/2017/file/9ef2ed4b7fd2c810847ffa5fa85bce38-Paper.pdf} {Simple and scalable predictive uncertainty estimation using deep ensembles}.
\newblock In \emph{Advances in Neural Information Processing Systems}, volume~30. Curran Associates, Inc.

\bibitem[{Liang et~al.(2022)Liang, Bommasani, Lee, Tsipras, Soylu, Yasunaga, Zhang, Narayanan, Wu, Kumar et~al.}]{liang2022holistic}
Percy Liang, Rishi Bommasani, Tony Lee, Dimitris Tsipras, Dilara Soylu, Michihiro Yasunaga, Yian Zhang, Deepak Narayanan, Yuhuai Wu, Ananya Kumar, et~al. 2022.
\newblock Holistic evaluation of language models.
\newblock \emph{arXiv preprint arXiv:2211.09110}.

\bibitem[{Liang et~al.(2023)Liang, He, Jiao, Wang, Wang, Wang, Yang, Tu, and Shi}]{liang2023encouraging}
Tian Liang, Zhiwei He, Wenxiang Jiao, Xing Wang, Yan Wang, Rui Wang, Yujiu Yang, Zhaopeng Tu, and Shuming Shi. 2023.
\newblock Encouraging divergent thinking in large language models through multi-agent debate.
\newblock \emph{arXiv preprint arXiv:2305.19118}.

\bibitem[{OpenAI(2024)}]{openai_gpt4o_2024}
OpenAI. 2024.
\newblock Introducing gpt-4o.
\newblock \url{https://openai.com/index/hello-gpt-4o/}.
\newblock Accessed: 2024-09-16.

\bibitem[{Ovadia et~al.(2019)Ovadia, Fertig, Ren, Nado, Sculley, Nowozin, Dillon, Lakshminarayanan, and Snoek}]{ovadia2019can}
Yaniv Ovadia, Emily Fertig, Jie Ren, Zachary Nado, David Sculley, Sebastian Nowozin, Joshua Dillon, Balaji Lakshminarayanan, and Jasper Snoek. 2019.
\newblock Can you trust your model's uncertainty? evaluating predictive uncertainty under dataset shift.
\newblock \emph{Advances in neural information processing systems}, 32.

\bibitem[{Rae et~al.(2021)Rae, Borgeaud, Cai, Millican, Hoffmann, Song, Aslanides, Henderson, Ring, Young et~al.}]{rae2021scaling}
Jack~W Rae, Sebastian Borgeaud, Trevor Cai, Katie Millican, Jordan Hoffmann, Francis Song, John Aslanides, Sarah Henderson, Roman Ring, Susannah Young, et~al. 2021.
\newblock Scaling language models: Methods, analysis \& insights from training gopher.
\newblock \emph{arXiv preprint arXiv:2112.11446}.

\bibitem[{Reimers and Gurevych(2019)}]{reimers-gurevych-2019-sentence}
Nils Reimers and Iryna Gurevych. 2019.
\newblock \href {https://doi.org/10.18653/v1/D19-1410} {Sentence-{BERT}: Sentence embeddings using {S}iamese {BERT}-networks}.
\newblock In \emph{Proceedings of the 2019 Conference on Empirical Methods in Natural Language Processing and the 9th International Joint Conference on Natural Language Processing (EMNLP-IJCNLP)}, pages 3982--3992, Hong Kong, China. Association for Computational Linguistics.

\bibitem[{Ribeiro et~al.(2020)Ribeiro, Wu, Guestrin, and Singh}]{ribeiro2020beyond}
Marco~Tulio Ribeiro, Tongshuang Wu, Carlos Guestrin, and Sameer Singh. 2020.
\newblock Beyond accuracy: Behavioral testing of nlp models with checklist.
\newblock \emph{arXiv preprint arXiv:2005.04118}.

\bibitem[{Rotman and Reichart(2022)}]{rotman2022multi}
Guy Rotman and Roi Reichart. 2022.
\newblock Multi-task active learning for pre-trained transformer-based models.
\newblock \emph{Transactions of the Association for Computational Linguistics}, 10:1209--1228.

\bibitem[{Settles and Craven(2008)}]{settles2008analysis}
Burr Settles and Mark Craven. 2008.
\newblock An analysis of active learning strategies for sequence labeling tasks.
\newblock In \emph{Proceedings of the 2008 Conference on Empirical Methods in Natural Language Processing}, pages 1070--1079.

\bibitem[{Shum et~al.(2023)Shum, Diao, and Zhang}]{shum-etal-2023-automatic}
Kashun Shum, Shizhe Diao, and Tong Zhang. 2023.
\newblock \href {https://doi.org/10.18653/v1/2023.findings-emnlp.811} {Automatic prompt augmentation and selection with chain-of-thought from labeled data}.
\newblock In \emph{Findings of the Association for Computational Linguistics: EMNLP 2023}, pages 12113--12139, Singapore. Association for Computational Linguistics.

\bibitem[{Sileo and Lernould(2023)}]{sileo2023mindgames}
Damien Sileo and Antoine Lernould. 2023.
\newblock Mindgames: Targeting theory of mind in large language models with dynamic epistemic modal logic.
\newblock \emph{arXiv preprint arXiv:2305.03353}.

\bibitem[{Thoppilan et~al.(2022)Thoppilan, De~Freitas, Hall, Shazeer, Kulshreshtha, Cheng, Jin, Bos, Baker, Du et~al.}]{thoppilan2022lamda}
Romal Thoppilan, Daniel De~Freitas, Jamie Hall, Noam Shazeer, Apoorv Kulshreshtha, Heng-Tze Cheng, Alicia Jin, Taylor Bos, Leslie Baker, Yu~Du, et~al. 2022.
\newblock Lamda: Language models for dialog applications.
\newblock \emph{arXiv preprint arXiv:2201.08239}.

\bibitem[{Tomani et~al.(2024)Tomani, Chaudhuri, Evtimov, Cremers, and Ibrahim}]{tomani2024uncertainty}
Christian Tomani, Kamalika Chaudhuri, Ivan Evtimov, Daniel Cremers, and Mark Ibrahim. 2024.
\newblock Uncertainty-based abstention in llms improves safety and reduces hallucinations.
\newblock \emph{arXiv preprint arXiv:2404.10960}.

\bibitem[{Touvron et~al.(2023)Touvron, Lavril, Izacard, Martinet, Lachaux, Lacroix, Rozi{\`e}re, Goyal, Hambro, Azhar et~al.}]{touvron2023llama}
Hugo Touvron, Thibaut Lavril, Gautier Izacard, Xavier Martinet, Marie-Anne Lachaux, Timoth{\'e}e Lacroix, Baptiste Rozi{\`e}re, Naman Goyal, Eric Hambro, Faisal Azhar, et~al. 2023.
\newblock Llama: Open and efficient foundation language models.
\newblock \emph{arXiv preprint arXiv:2302.13971}.

\bibitem[{Van~Amersfoort et~al.(2020)Van~Amersfoort, Smith, Teh, and Gal}]{pmlr-v119-van-amersfoort20a}
Joost Van~Amersfoort, Lewis Smith, Yee~Whye Teh, and Yarin Gal. 2020.
\newblock \href {https://proceedings.mlr.press/v119/van-amersfoort20a.html} {Uncertainty estimation using a single deep deterministic neural network}.
\newblock In \emph{Proceedings of the 37th International Conference on Machine Learning}, volume 119 of \emph{Proceedings of Machine Learning Research}, pages 9690--9700. PMLR.

\bibitem[{Vashurin et~al.(2024)Vashurin, Fadeeva, Vazhentsev, Tsvigun, Vasilev, Xing, Sadallah, Rvanova, Petrakov, Panchenko et~al.}]{vashurin2024benchmarking}
Roman Vashurin, Ekaterina Fadeeva, Artem Vazhentsev, Akim Tsvigun, Daniil Vasilev, Rui Xing, Abdelrahman~Boda Sadallah, Lyudmila Rvanova, Sergey Petrakov, Alexander Panchenko, et~al. 2024.
\newblock Benchmarking uncertainty quantification methods for large language models with lm-polygraph.
\newblock \emph{arXiv preprint arXiv:2406.15627}.

\bibitem[{Wan et~al.(2023)Wan, Sun, Dai, Arik, and Pfister}]{wan-etal-2023-better}
Xingchen Wan, Ruoxi Sun, Hanjun Dai, Sercan Arik, and Tomas Pfister. 2023.
\newblock \href {https://doi.org/10.18653/v1/2023.findings-acl.216} {Better zero-shot reasoning with self-adaptive prompting}.
\newblock In \emph{Findings of the Association for Computational Linguistics: ACL 2023}, pages 3493--3514, Toronto, Canada. Association for Computational Linguistics.

\bibitem[{Wang et~al.(2022)Wang, Wei, Schuurmans, Le, Chi, Narang, Chowdhery, and Zhou}]{wang2022self}
Xuezhi Wang, Jason Wei, Dale Schuurmans, Quoc~V Le, Ed~H Chi, Sharan Narang, Aakanksha Chowdhery, and Denny Zhou. 2022.
\newblock Self-consistency improves chain of thought reasoning in language models.
\newblock In \emph{The Eleventh International Conference on Learning Representations}.

\bibitem[{Wang et~al.(2023)Wang, Mao, Wu, Ge, Wei, and Ji}]{wang2023unleashing}
Zhenhailong Wang, Shaoguang Mao, Wenshan Wu, Tao Ge, Furu Wei, and Heng Ji. 2023.
\newblock Unleashing cognitive synergy in large language models: A task-solving agent through multi-persona selfcollaboration.
\newblock \emph{arXiv preprint arXiv:2307.05300}, 1(2):3.

\bibitem[{Wei et~al.(2022)Wei, Wang, Schuurmans, Bosma, Xia, Chi, Le, Zhou et~al.}]{wei2022chain}
Jason Wei, Xuezhi Wang, Dale Schuurmans, Maarten Bosma, Fei Xia, Ed~Chi, Quoc~V Le, Denny Zhou, et~al. 2022.
\newblock Chain-of-thought prompting elicits reasoning in large language models.
\newblock \emph{Advances in Neural Information Processing Systems}, 35:24824--24837.

\bibitem[{Yao et~al.(2023)Yao, Yu, Zhao, Shafran, Griffiths, Cao, and Narasimhan}]{yao2023tree}
Shunyu Yao, Dian Yu, Jeffrey Zhao, Izhak Shafran, Thomas~L Griffiths, Yuan Cao, and Karthik Narasimhan. 2023.
\newblock Tree of thoughts: Deliberate problem solving with large language models.
\newblock \emph{arXiv preprint arXiv:2305.10601}.

\bibitem[{Yin et~al.(2023)Yin, Sun, Chang, Guo, Dai, Huang, and Qiu}]{yin-etal-2023-exchange}
Zhangyue Yin, Qiushi Sun, Cheng Chang, Qipeng Guo, Junqi Dai, Xuanjing Huang, and Xipeng Qiu. 2023.
\newblock \href {https://doi.org/10.18653/v1/2023.emnlp-main.936} {Exchange-of-thought: Enhancing large language model capabilities through cross-model communication}.
\newblock In \emph{Proceedings of the 2023 Conference on Empirical Methods in Natural Language Processing}, pages 15135--15153, Singapore. Association for Computational Linguistics.

\bibitem[{Zhang et~al.(2022)Zhang, Zhang, Li, and Smola}]{zhang2022automatic}
Zhuosheng Zhang, Aston Zhang, Mu~Li, and Alex Smola. 2022.
\newblock Automatic chain of thought prompting in large language models.
\newblock In \emph{The Eleventh International Conference on Learning Representations}.

\bibitem[{Zhu et~al.(2023)Zhu, Wang, Zhang, Zhang, Huang, Gan, Zhang, and Yang}]{zhu-etal-2023-solving}
Xinyu Zhu, Junjie Wang, Lin Zhang, Yuxiang Zhang, Yongfeng Huang, Ruyi Gan, Jiaxing Zhang, and Yujiu Yang. 2023.
\newblock \href {https://doi.org/10.18653/v1/2023.acl-long.245} {Solving math word problems via cooperative reasoning induced language models}.
\newblock In \emph{Proceedings of the 61st Annual Meeting of the Association for Computational Linguistics (Volume 1: Long Papers)}, pages 4471--4485, Toronto, Canada. Association for Computational Linguistics.

\end{thebibliography}

\clearpage
\appendix
\section{Appendix} 
\begin{figure*}[!]
\begin{subfigure}{\textwidth}
\centering   \captionsetup{justification=centering} 
  \includegraphics[scale=0.15]{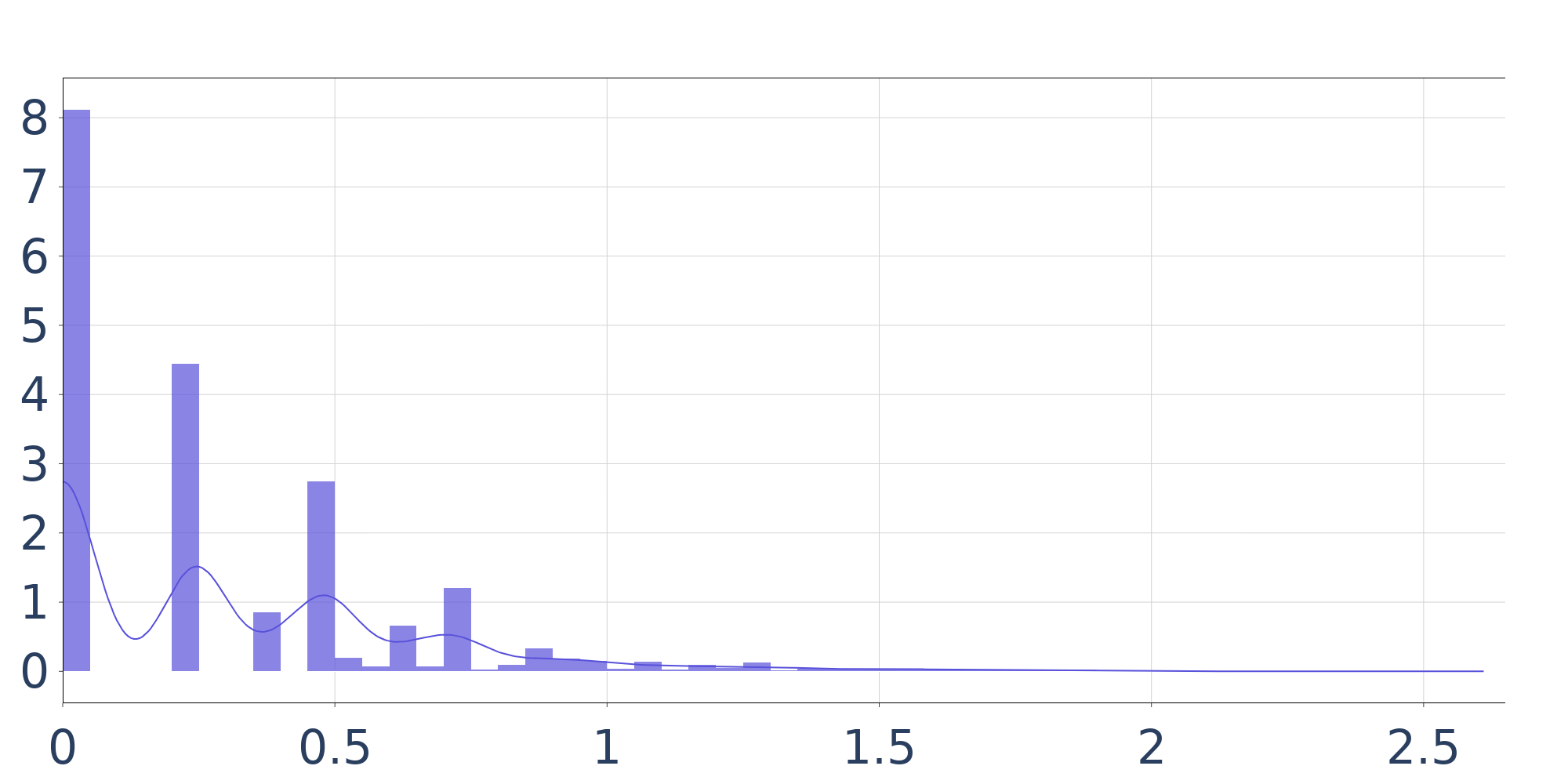}
  \caption{GSM8K}
\end{subfigure}
\hfill
\begin{subfigure}{\textwidth}
  \centering   \captionsetup{justification=centering} 
  \includegraphics[scale=0.15]{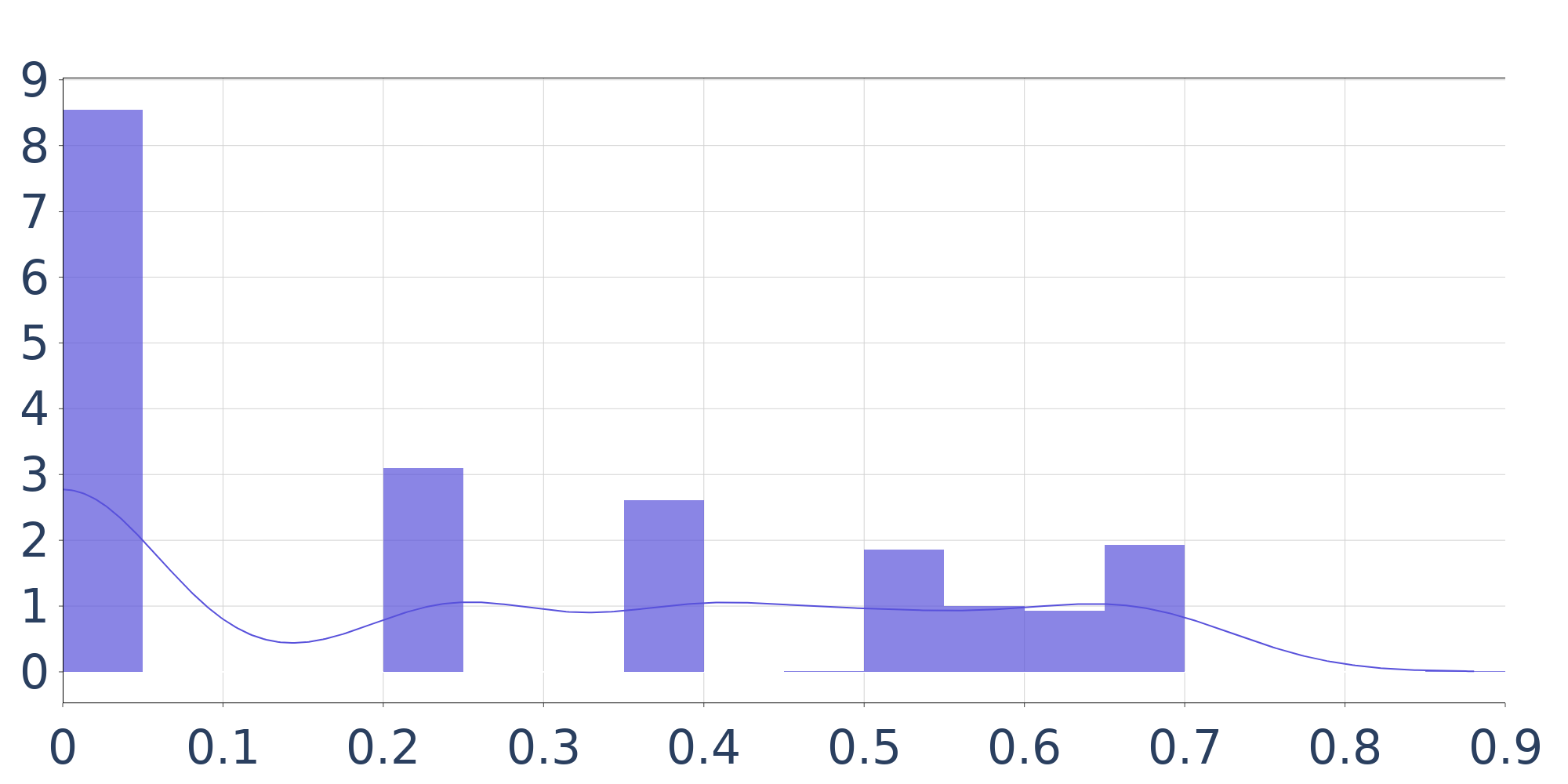}
  \caption{Fallacy}
\end{subfigure}
\hfill
\begin{subfigure}{\textwidth}
  \centering   \captionsetup{justification=centering} 
  \includegraphics[scale=0.15]{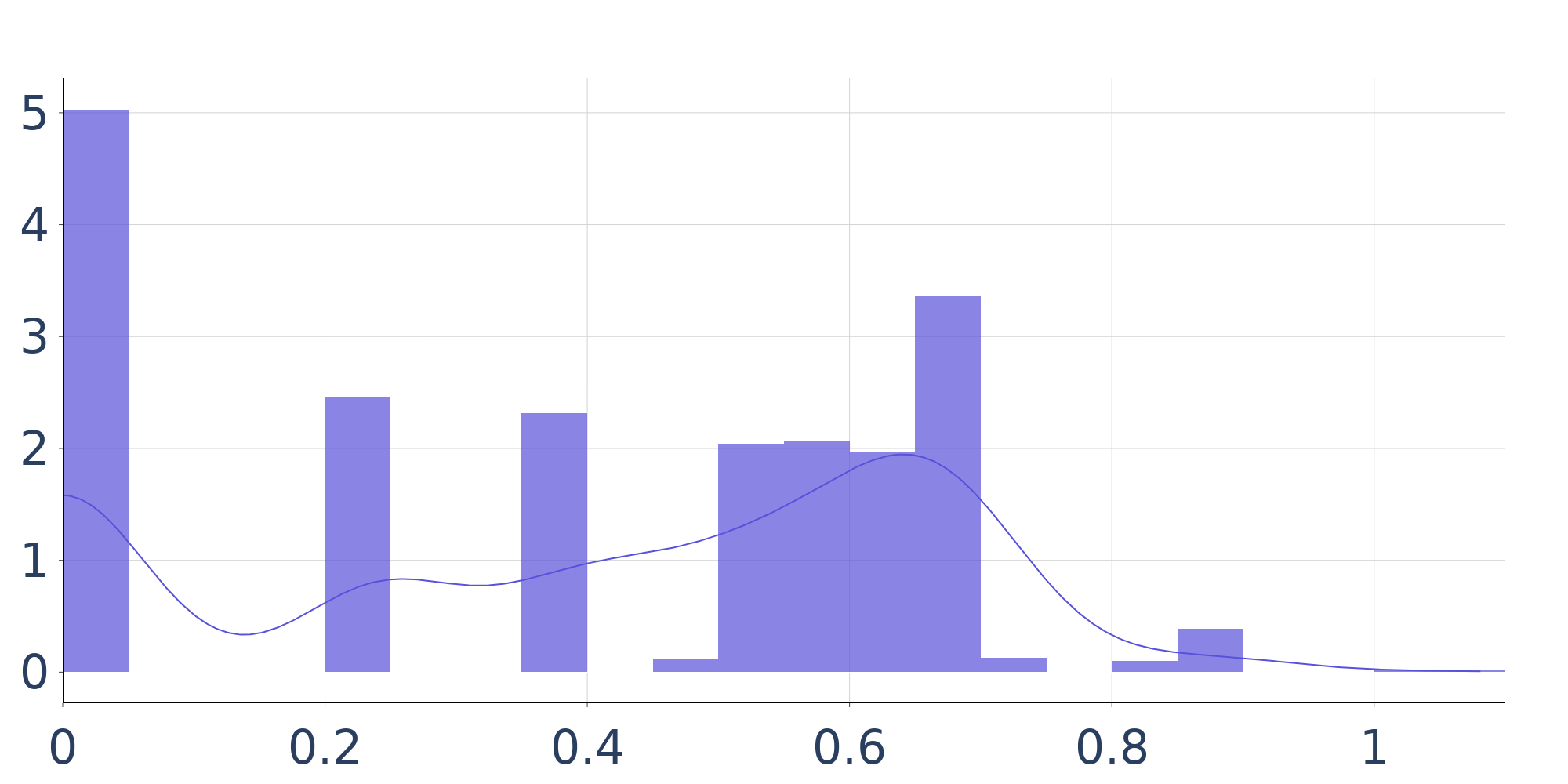}
  \caption{EPR}
\end{subfigure}
\hfill
\begin{subfigure}{\textwidth}
  \centering   \captionsetup{justification=centering} 
  \includegraphics[scale=0.15]{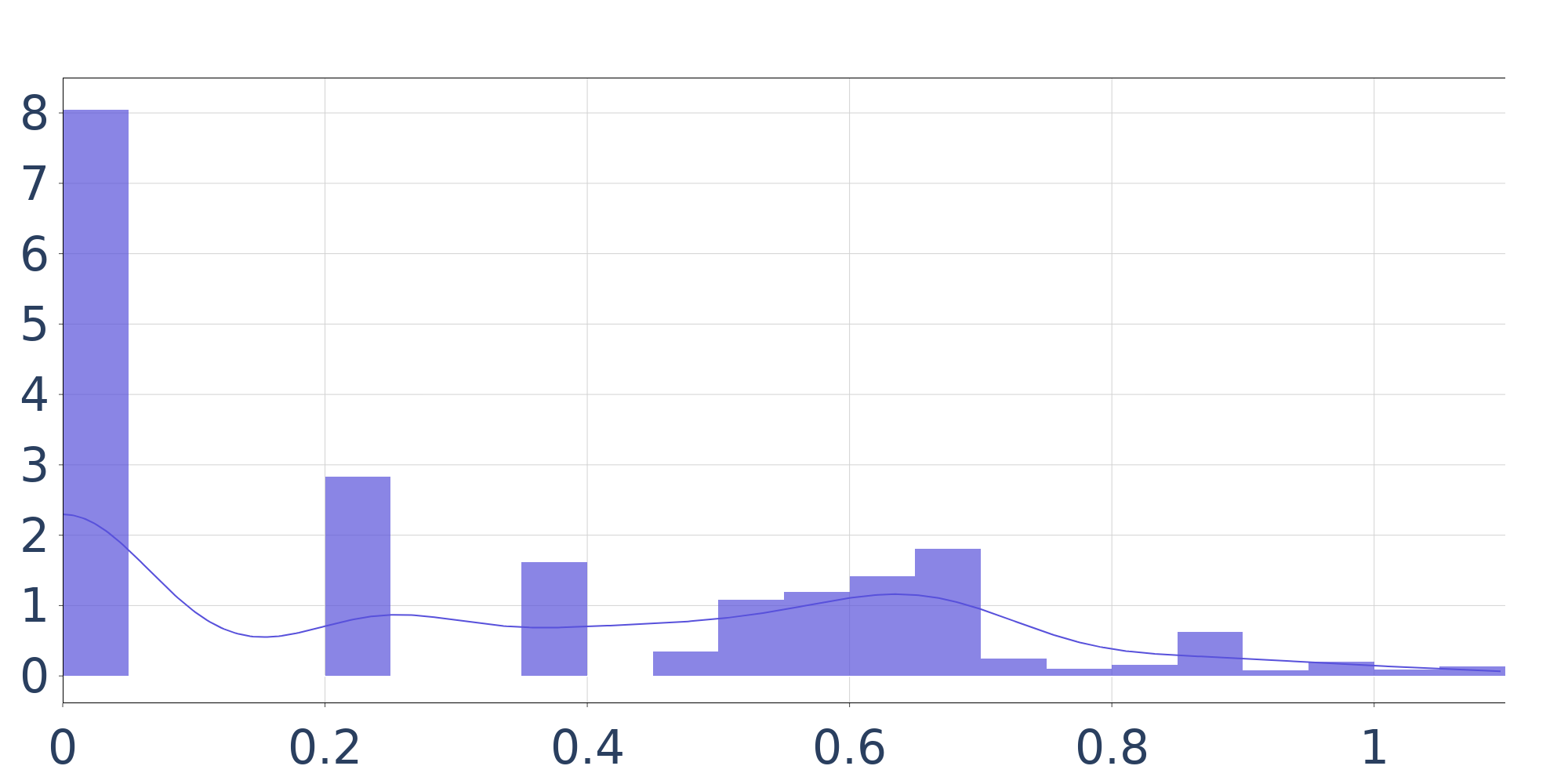}
  \caption{StrategyQA}
\end{subfigure}
\caption{Probability density function of uncertainty estimates of our method using \textbf{GPT4o}.}
\label{fig:uncertainty_gpt4o}
\end{figure*}

\begin{figure*}[!]
\begin{subfigure}{\textwidth}
\centering   \captionsetup{justification=centering} 
  \includegraphics[scale=0.15]{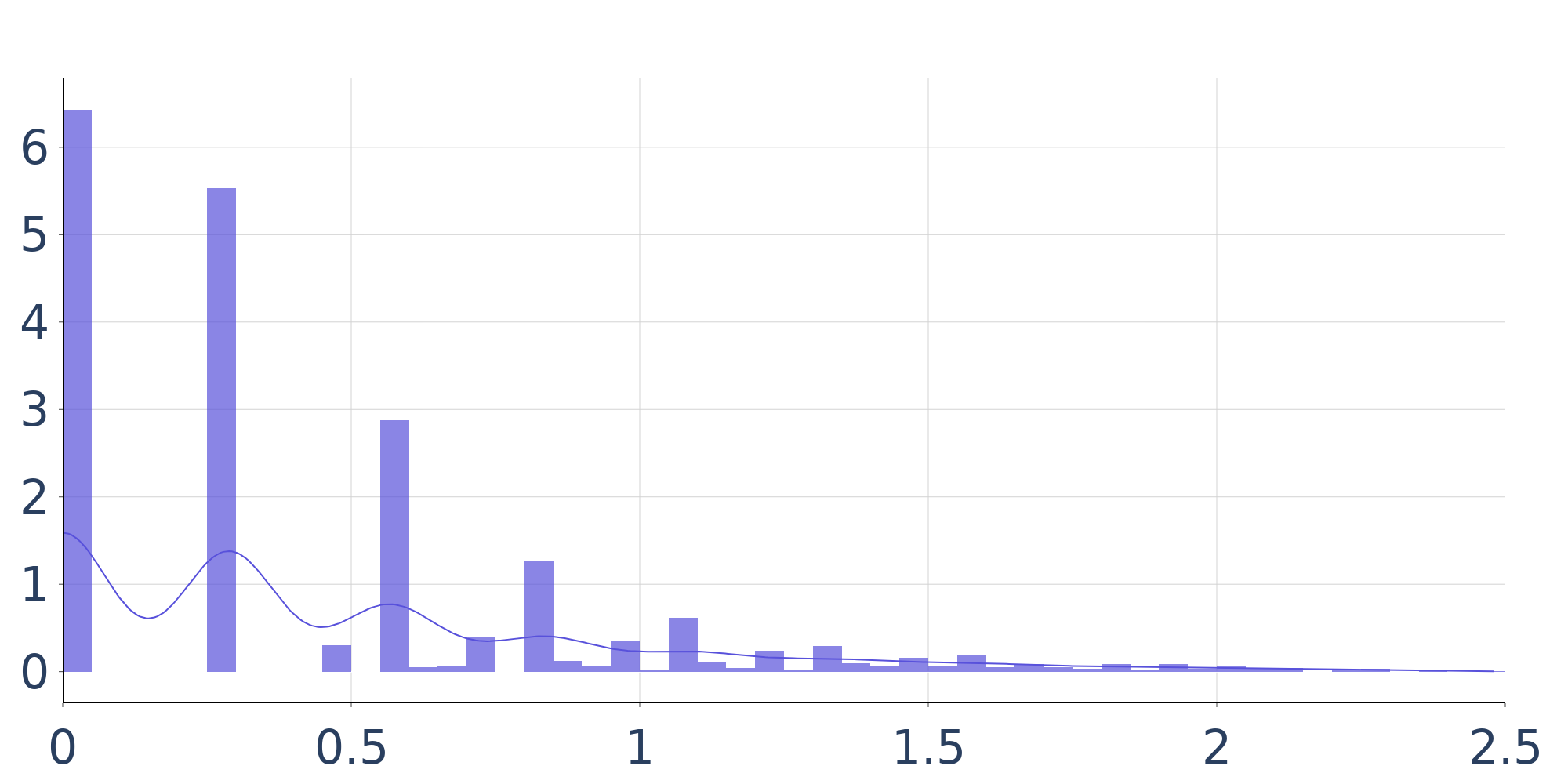}
  \caption{GSM8K}
\end{subfigure}
\hfill
\begin{subfigure}{\textwidth}
  \centering   \captionsetup{justification=centering} 
  \includegraphics[scale=0.15]{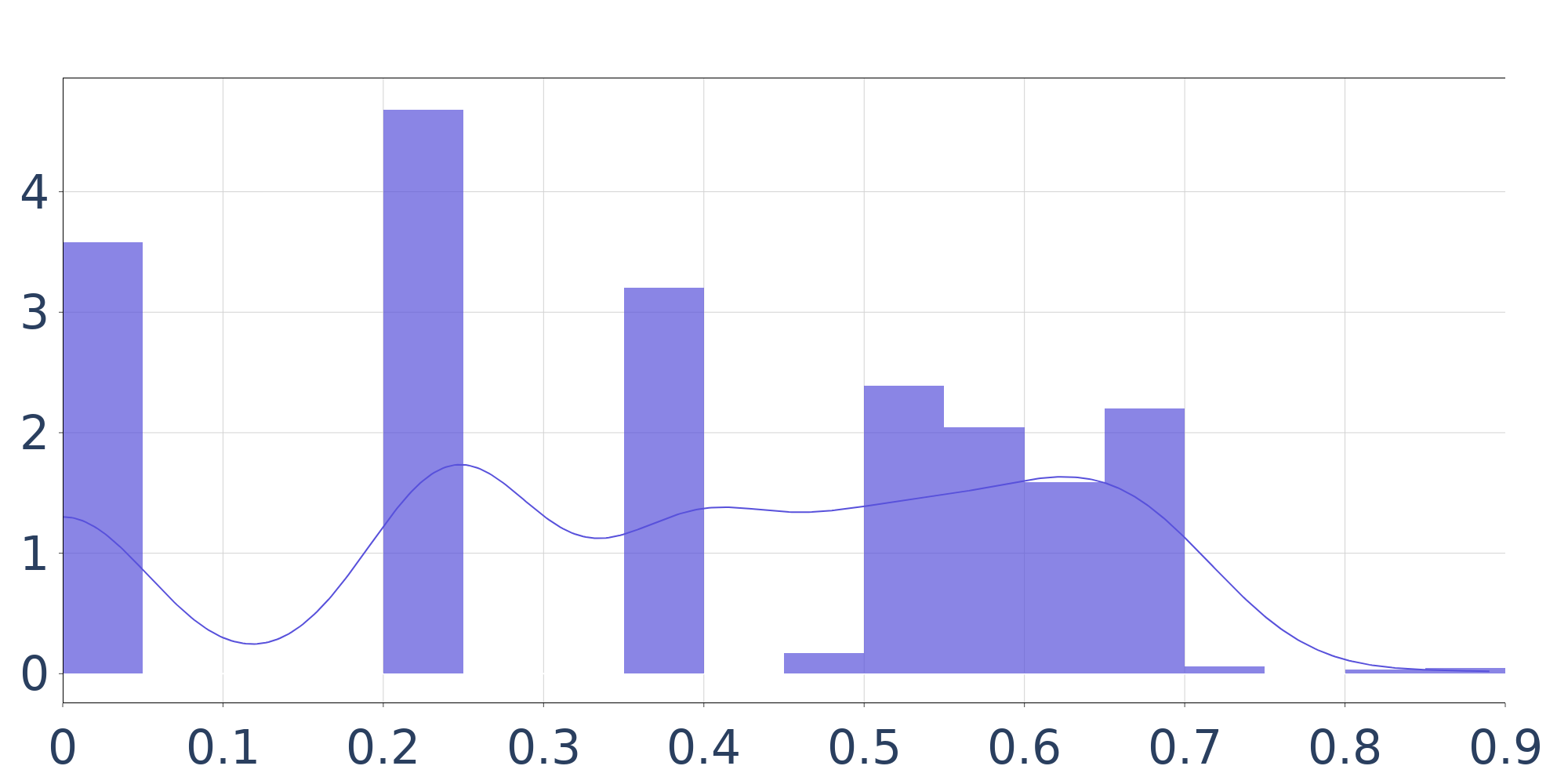}
  \caption{Fallacy}
\end{subfigure}
\hfill
\begin{subfigure}{\textwidth}
  \centering   \captionsetup{justification=centering} 
  \includegraphics[scale=0.15]{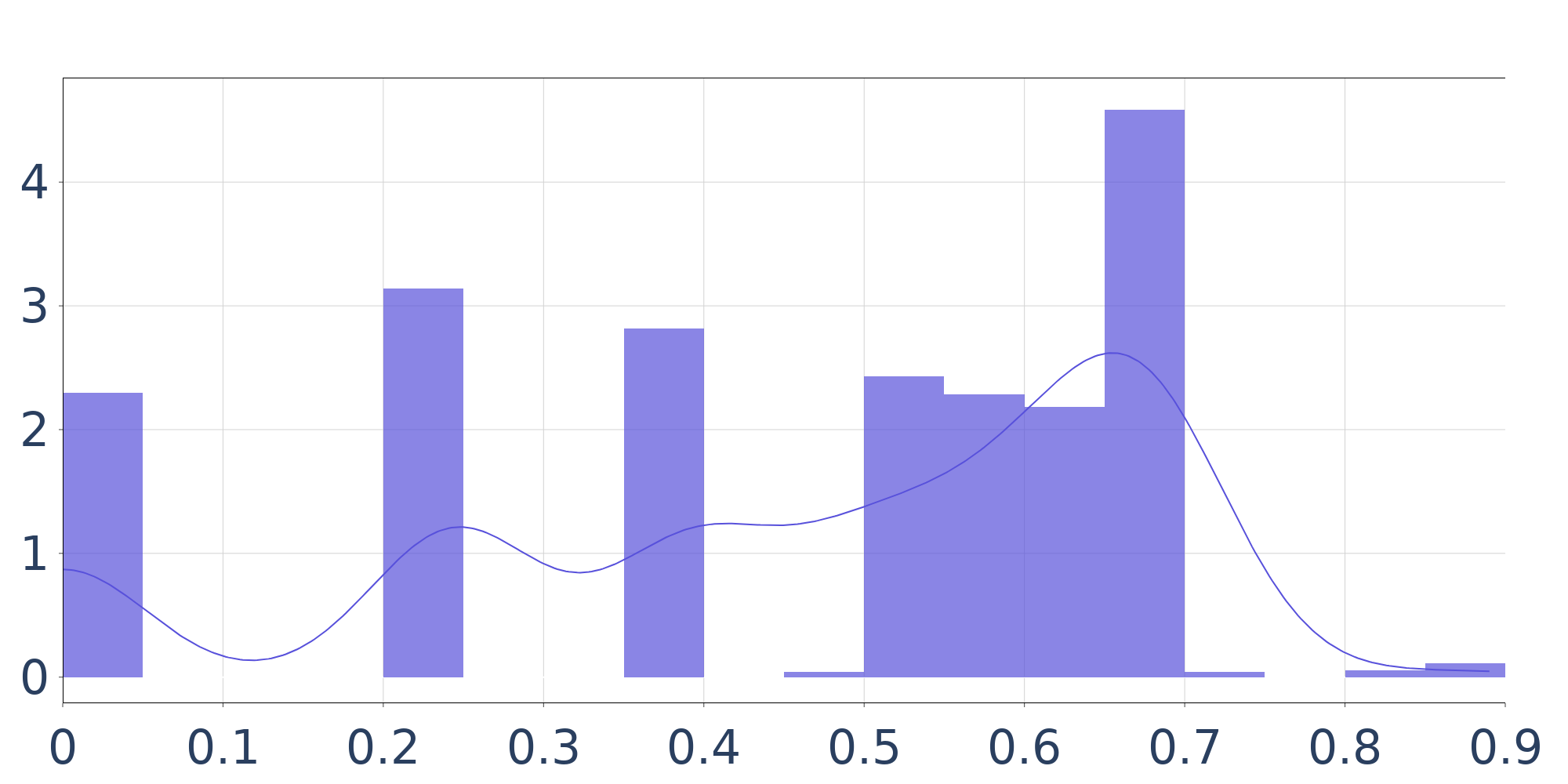}
  \caption{EPR}
\end{subfigure}
\hfill
\begin{subfigure}{\textwidth}
  \centering   \captionsetup{justification=centering} 
  \includegraphics[scale=0.15]{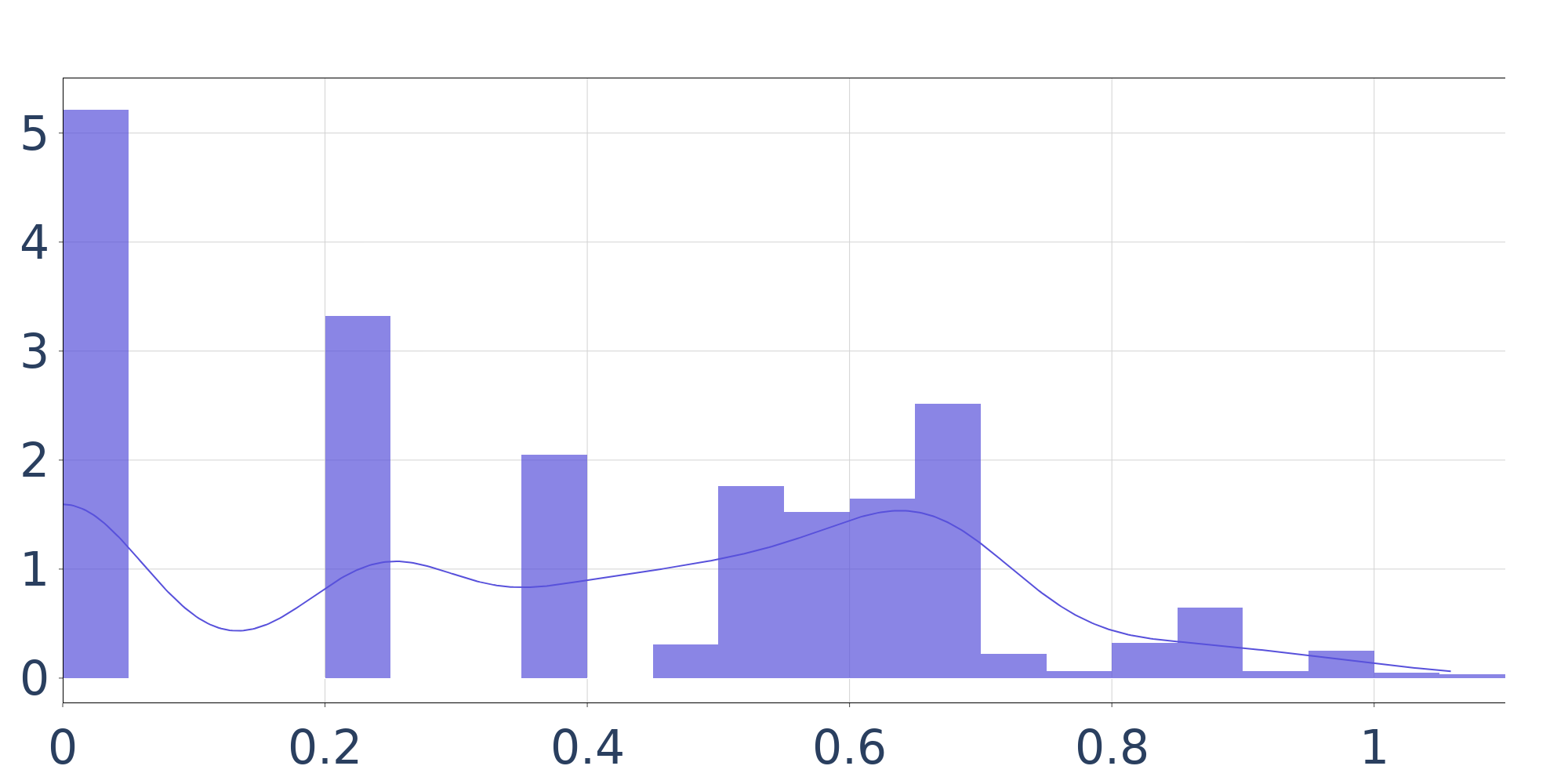}
  \caption{StrategyQA}
\end{subfigure}
\caption{Probability density function of uncertainty estimates of our method using \textbf{Phi3}.}
\label{fig:uncertainty_phi3}
\end{figure*}

\begin{figure*}[!]
\begin{subfigure}{\textwidth}
\centering   \captionsetup{justification=centering} 
  \includegraphics[scale=0.15]{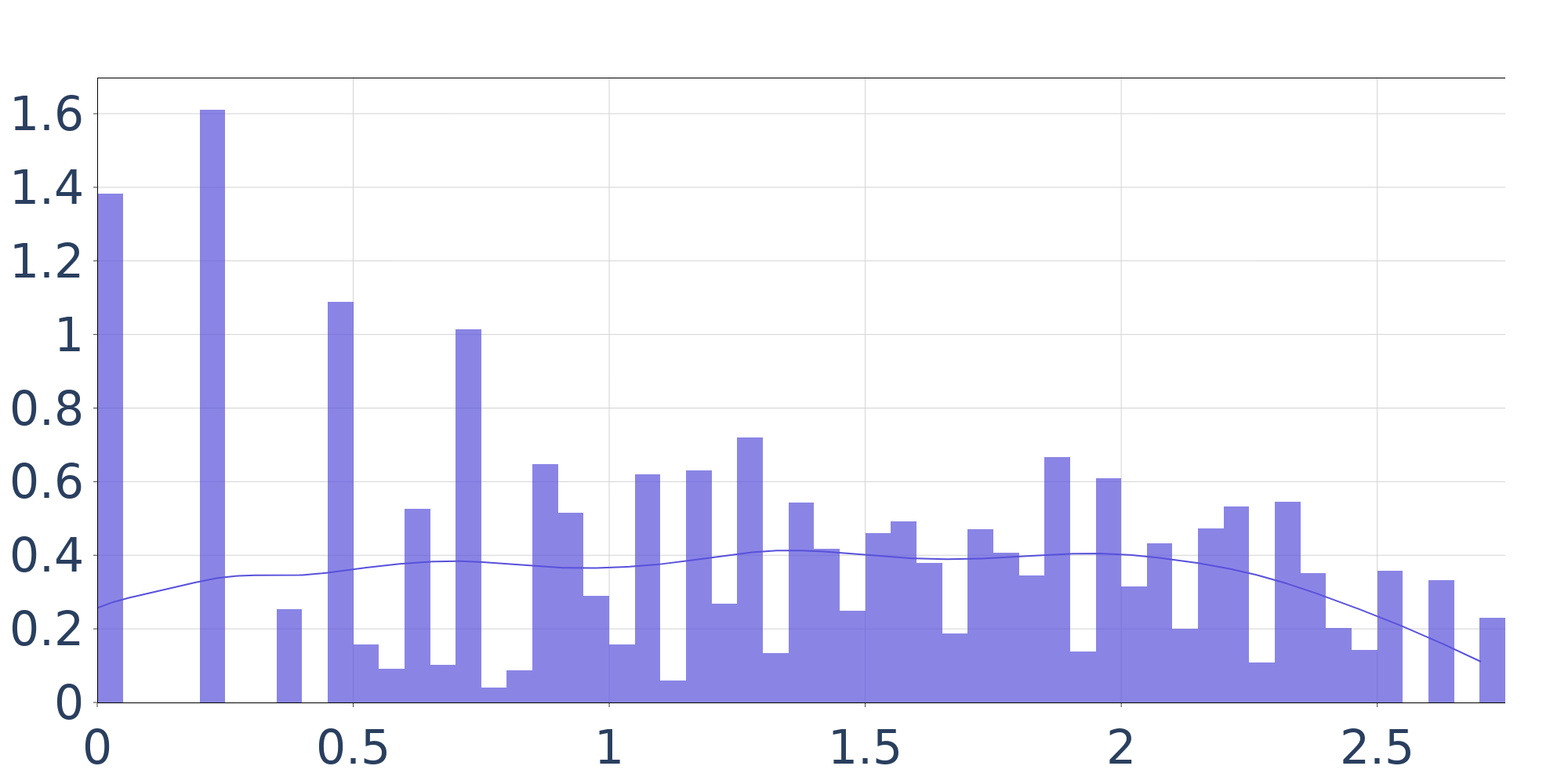}
  \caption{GSM8K}
\end{subfigure}
\hfill
\begin{subfigure}{\textwidth}
  \centering   \captionsetup{justification=centering} 
  \includegraphics[scale=0.15]{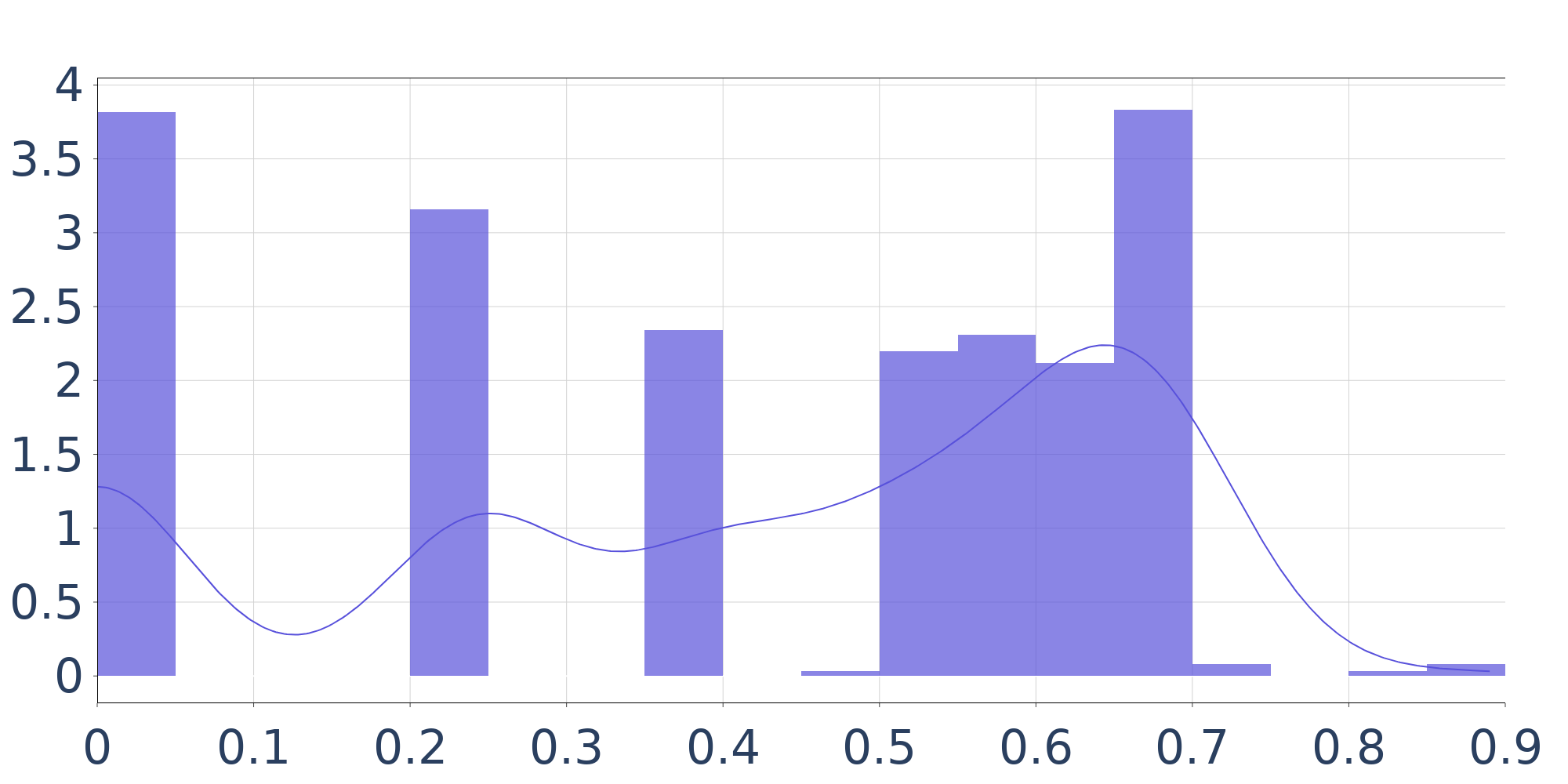}
  \caption{Fallacy}
\end{subfigure}
\hfill
\begin{subfigure}{\textwidth}
  \centering   \captionsetup{justification=centering} 
  \includegraphics[scale=0.15]{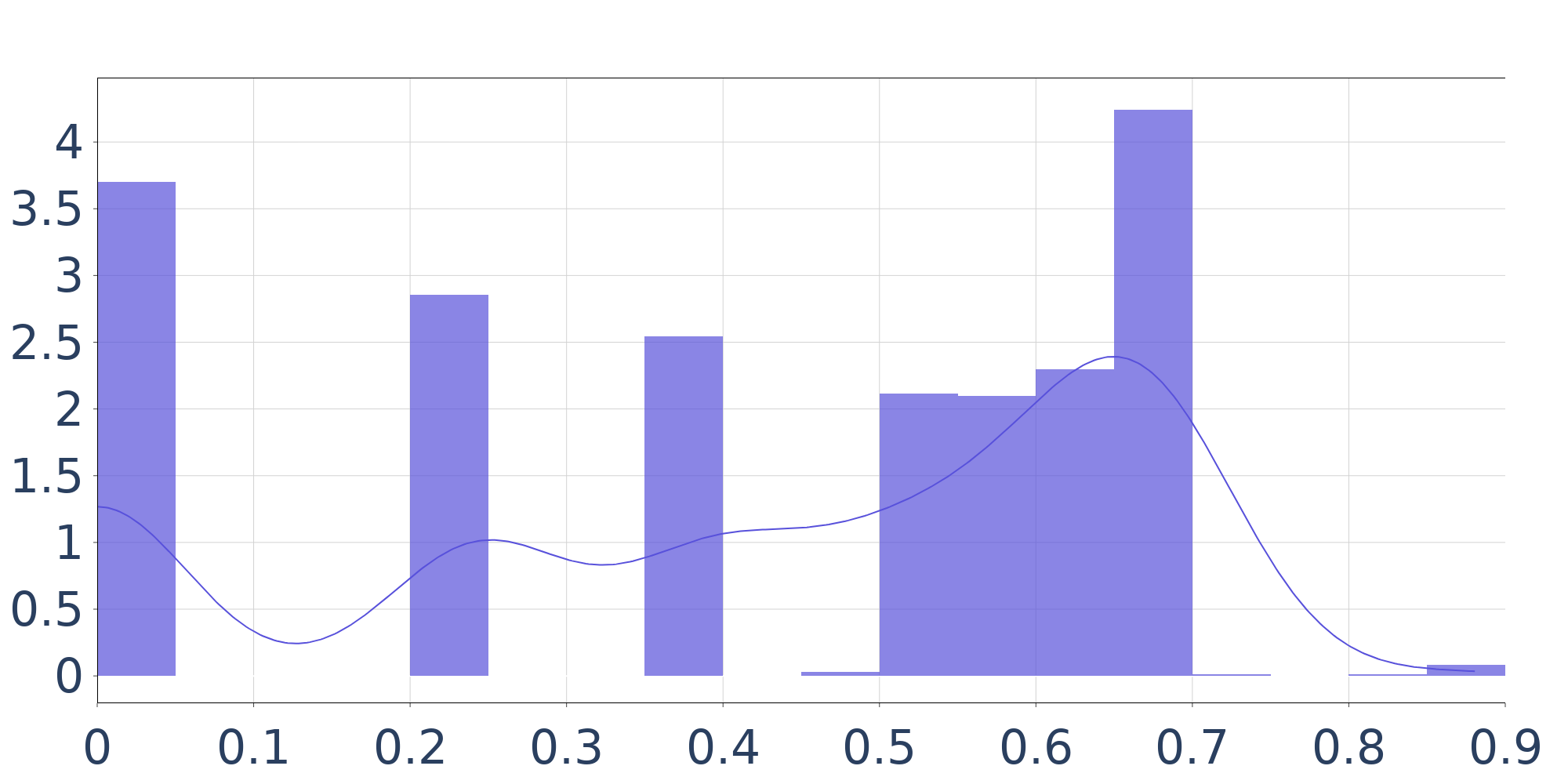}
  \caption{EPR}
\end{subfigure}
\hfill
\begin{subfigure}{\textwidth}
  \centering   \captionsetup{justification=centering} 
  \includegraphics[scale=0.15]{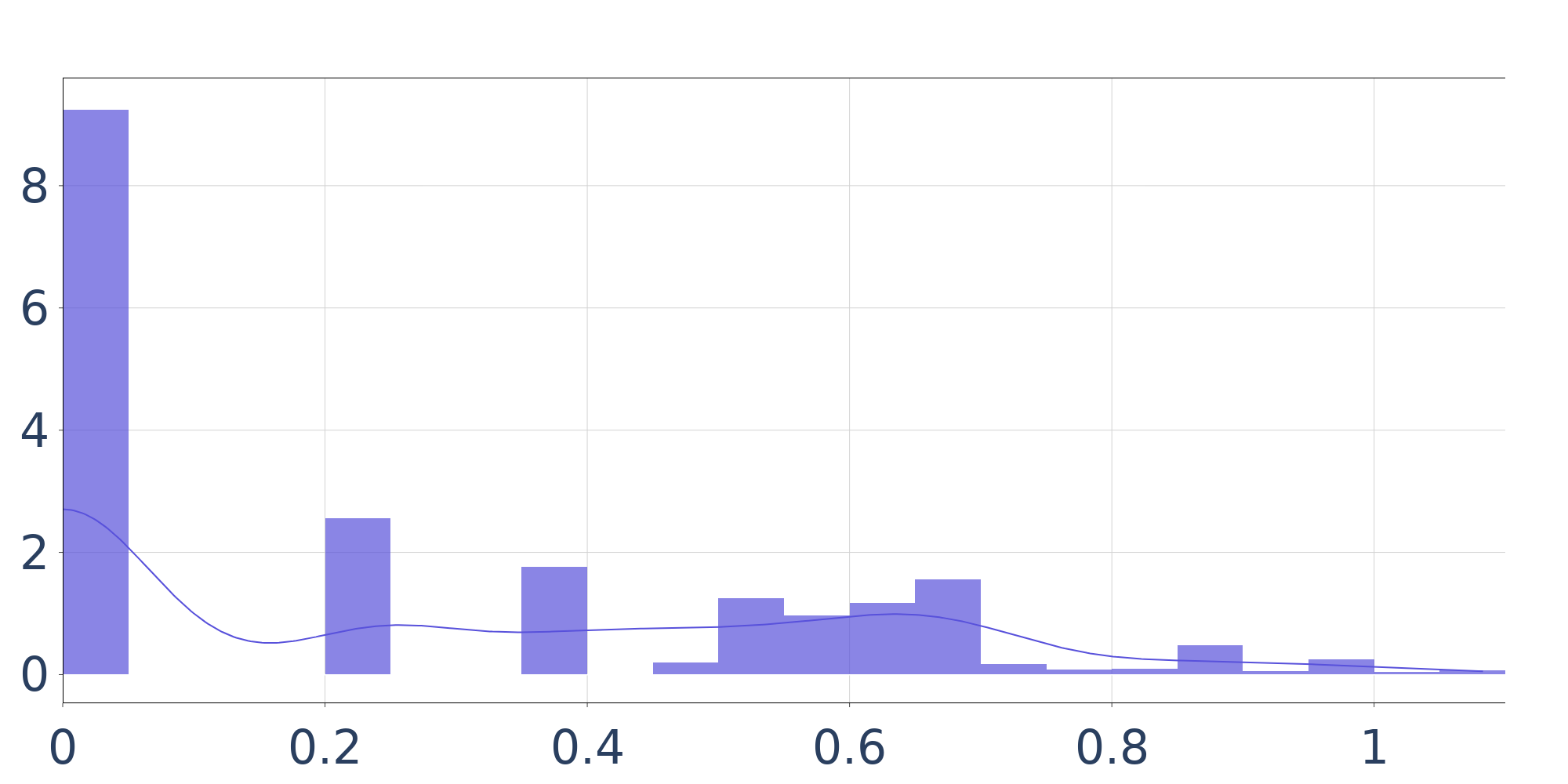}
  \caption{StrategyQA}
\end{subfigure}
\caption{Probability density function of uncertainty estimates of our method using \textbf{Mistral}.}
\label{fig:uncertainty_mistral}
\end{figure*}

\begin{figure*}[!]
\begin{subfigure}{\textwidth}
\centering   \captionsetup{justification=centering} 
  \includegraphics[scale=0.15]{figures/gsm8k_gpt3.5_uncertainty.png}
  \caption{GSM8K}
\end{subfigure}
\hfill
\begin{subfigure}{\textwidth}
  \centering   \captionsetup{justification=centering} 
  \includegraphics[scale=0.15]{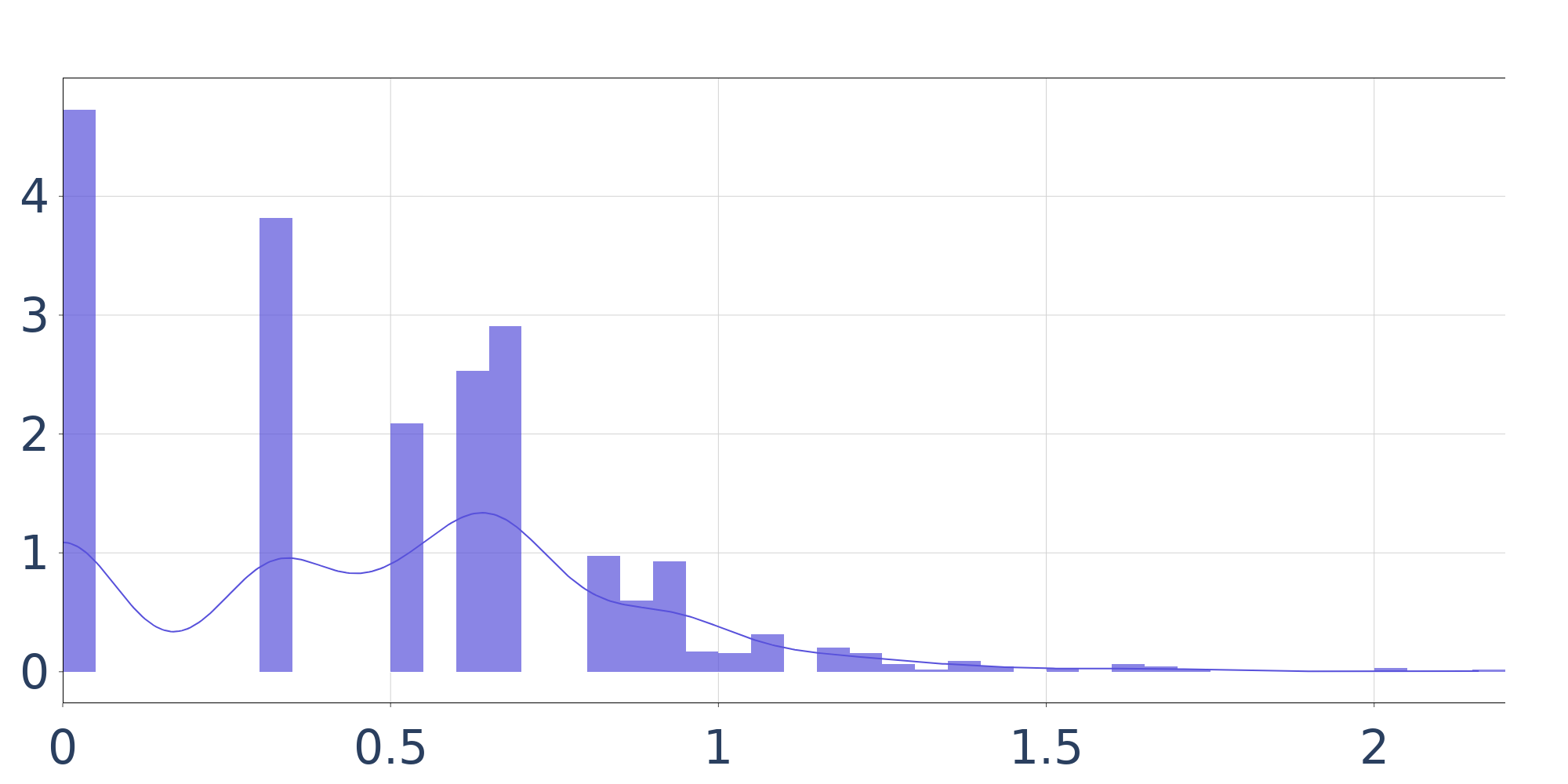}
  \caption{Fallacy}
\end{subfigure}
\hfill
\begin{subfigure}{\textwidth}
  \centering   \captionsetup{justification=centering} 
  \includegraphics[scale=0.15]{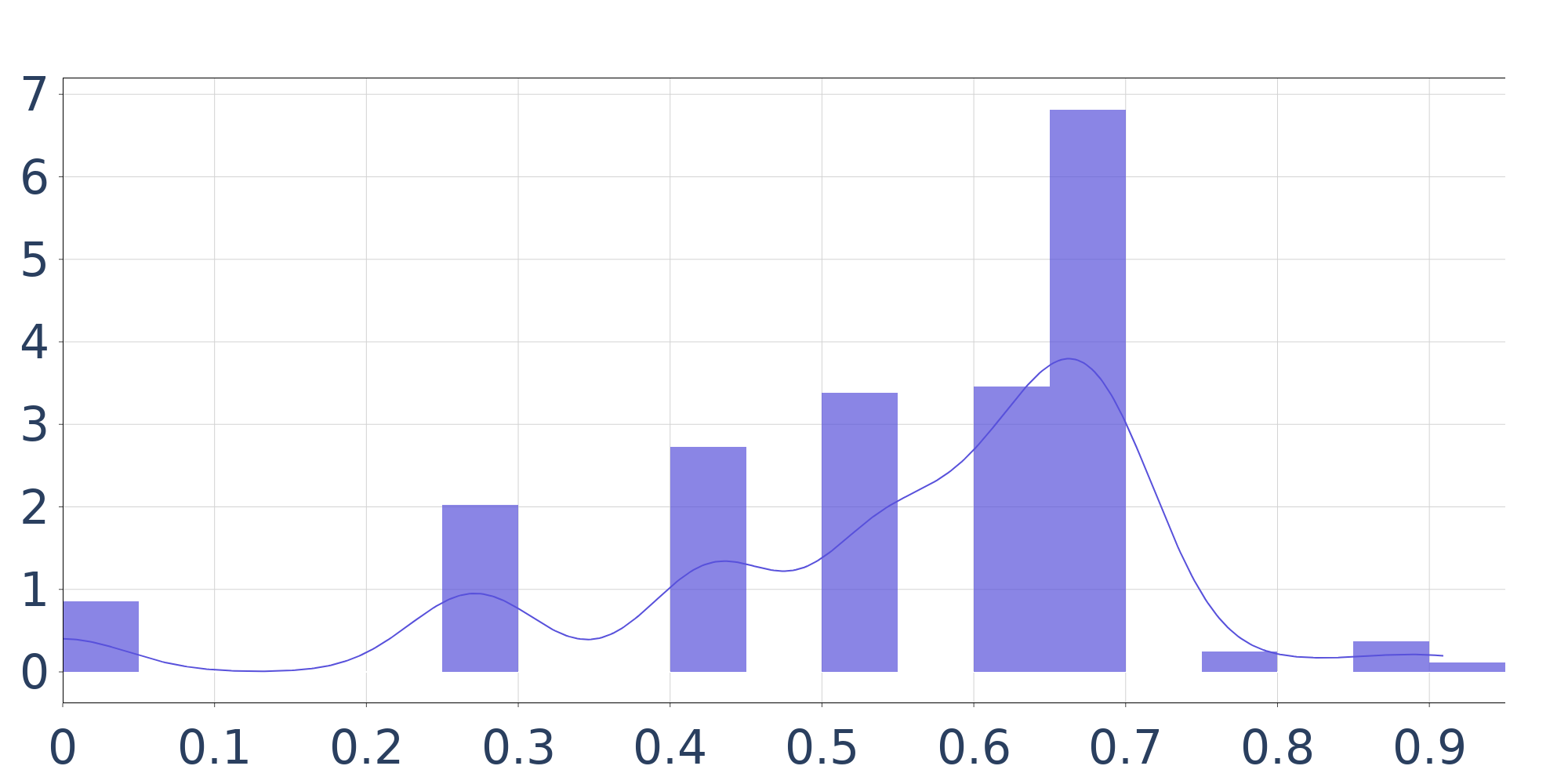}
  \caption{EPR}
\end{subfigure}
\hfill
\begin{subfigure}{\textwidth}
  \centering   \captionsetup{justification=centering} 
  \includegraphics[scale=0.15]{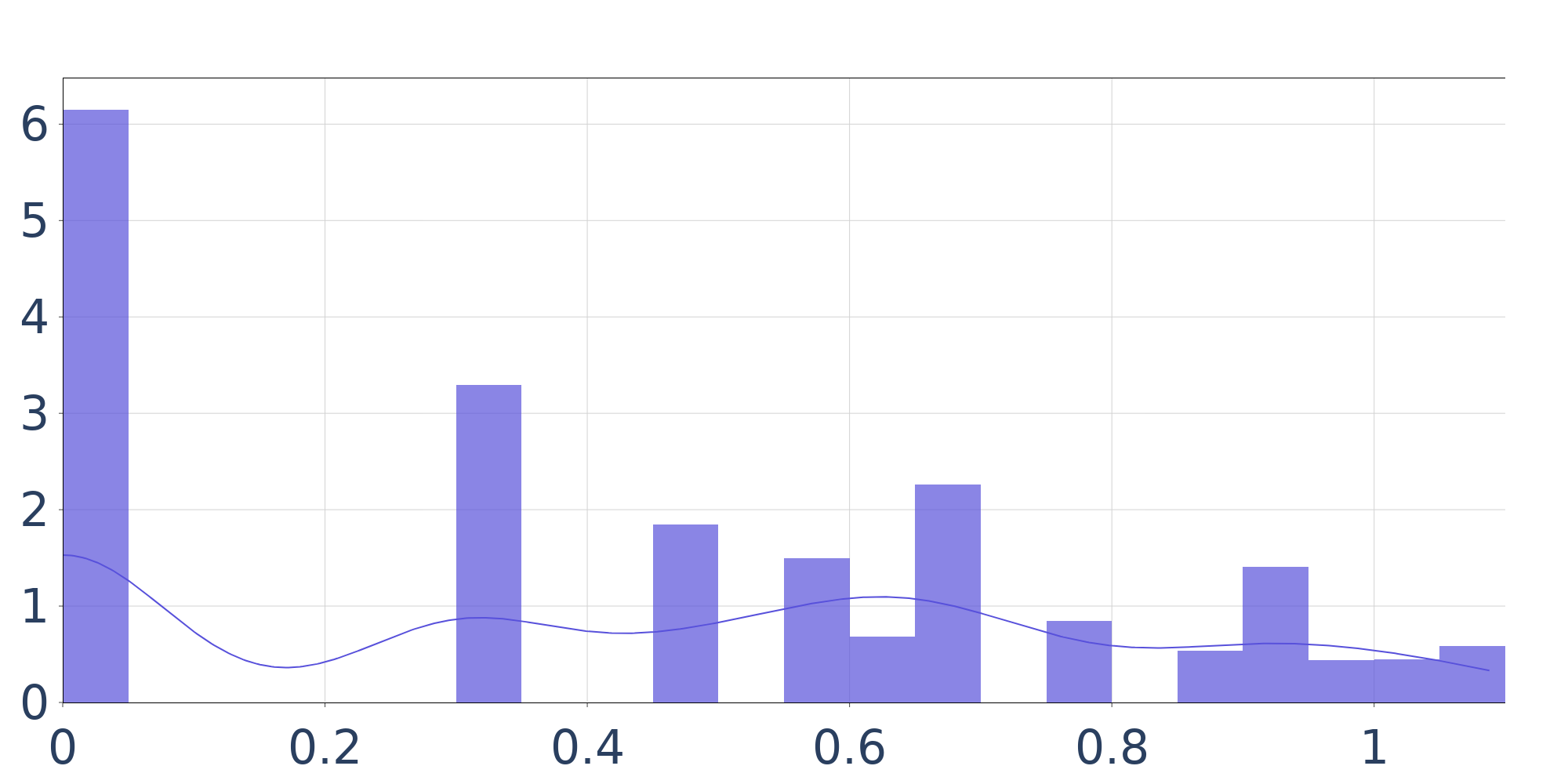}
  \caption{StrategyQA}
\end{subfigure}
\caption{Probability density function of uncertainty estimates of our method using \textbf{GPT3.5}.}
\label{fig:uncertainty_gpt3_5}
\end{figure*}

\begin{figure*}[!]
\begin{subfigure}{\textwidth}
\centering   \captionsetup{justification=centering} 
  \includegraphics[scale=0.15]{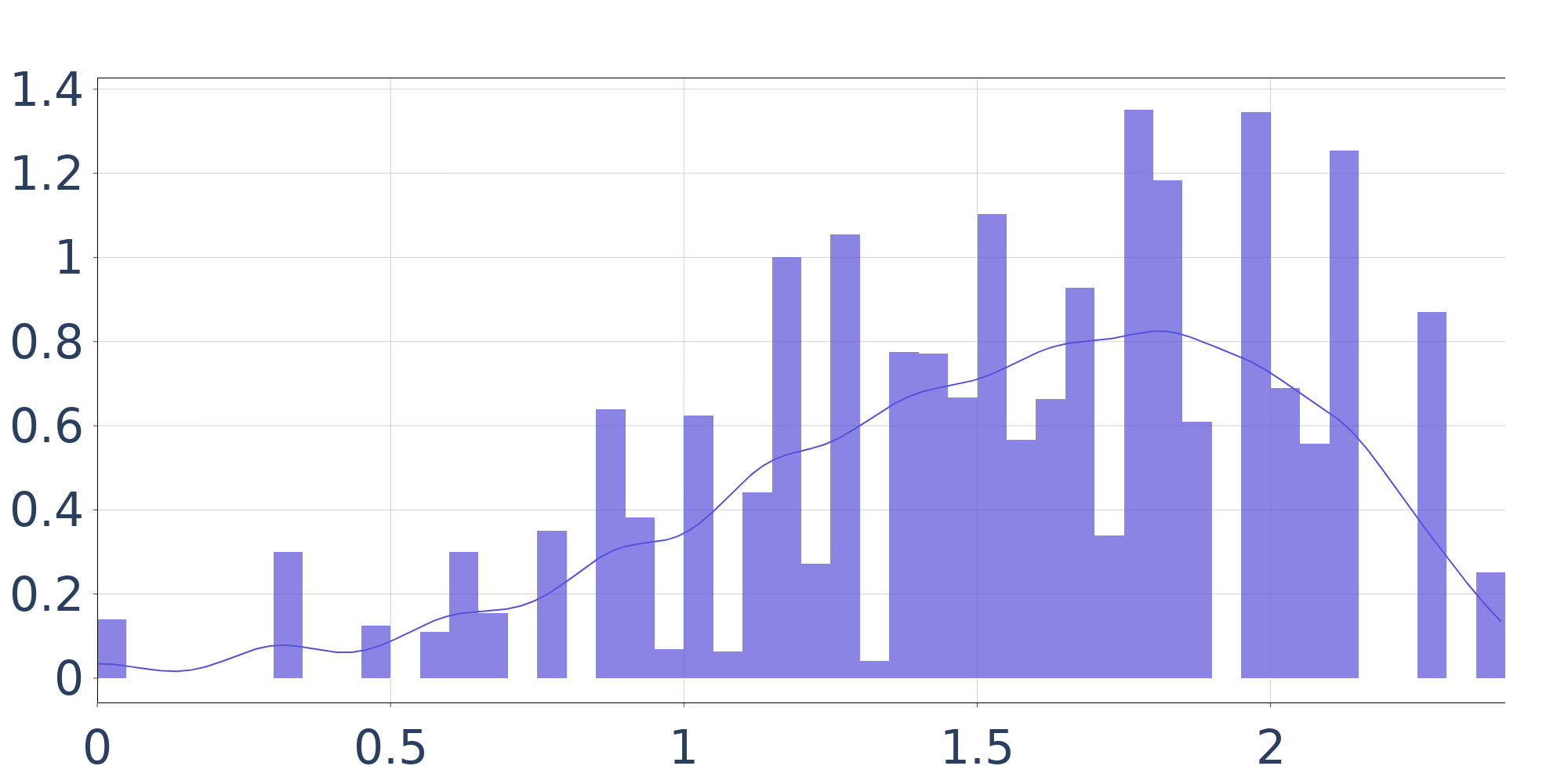}
  \caption{GSM8K}
\end{subfigure}
\hfill
\begin{subfigure}{\textwidth}
  \centering   \captionsetup{justification=centering} 
  \includegraphics[scale=0.15]{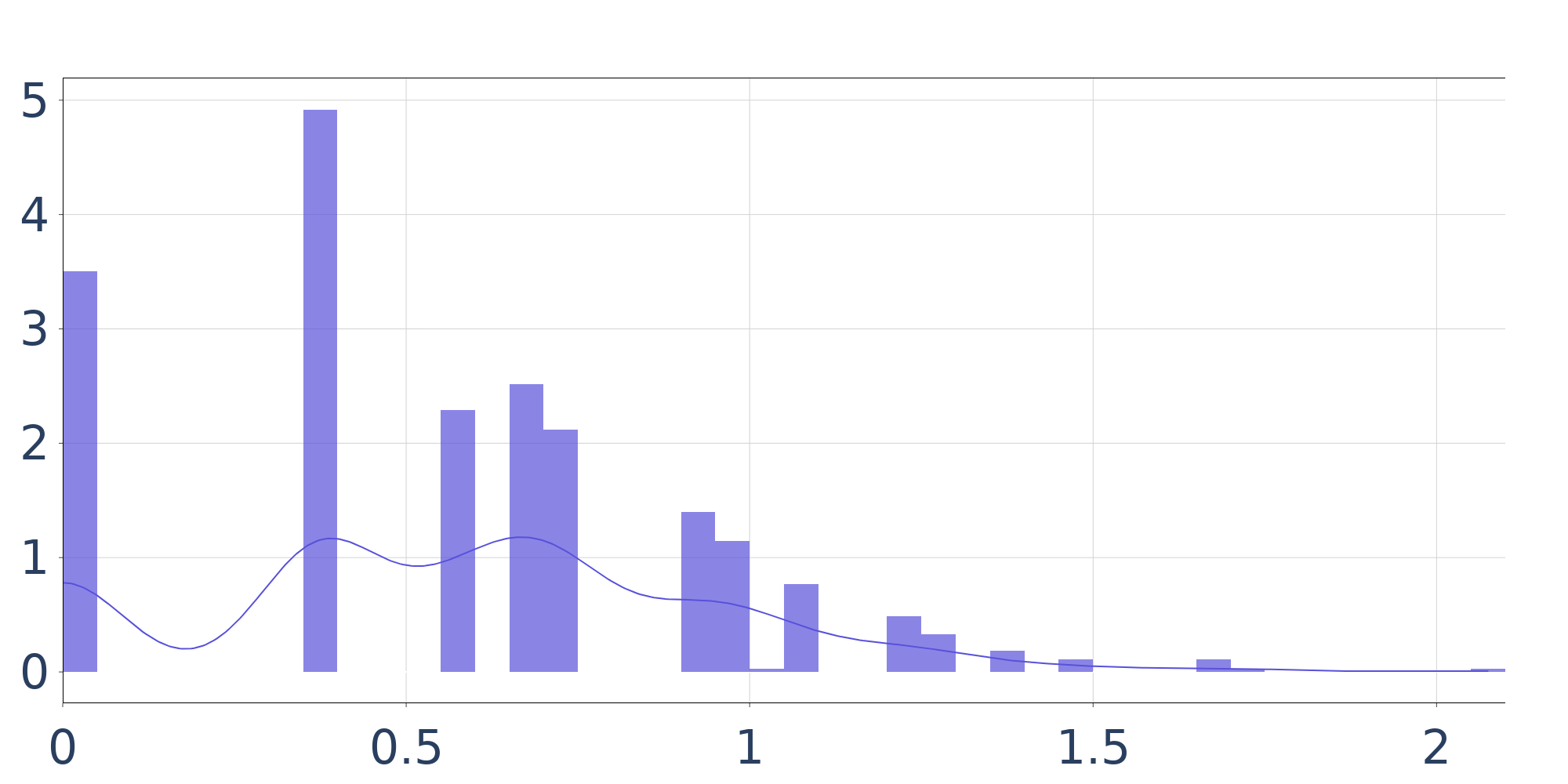}
  \caption{Fallacy}
\end{subfigure}
\hfill
\begin{subfigure}{\textwidth}
  \centering   \captionsetup{justification=centering} 
  \includegraphics[scale=0.15]{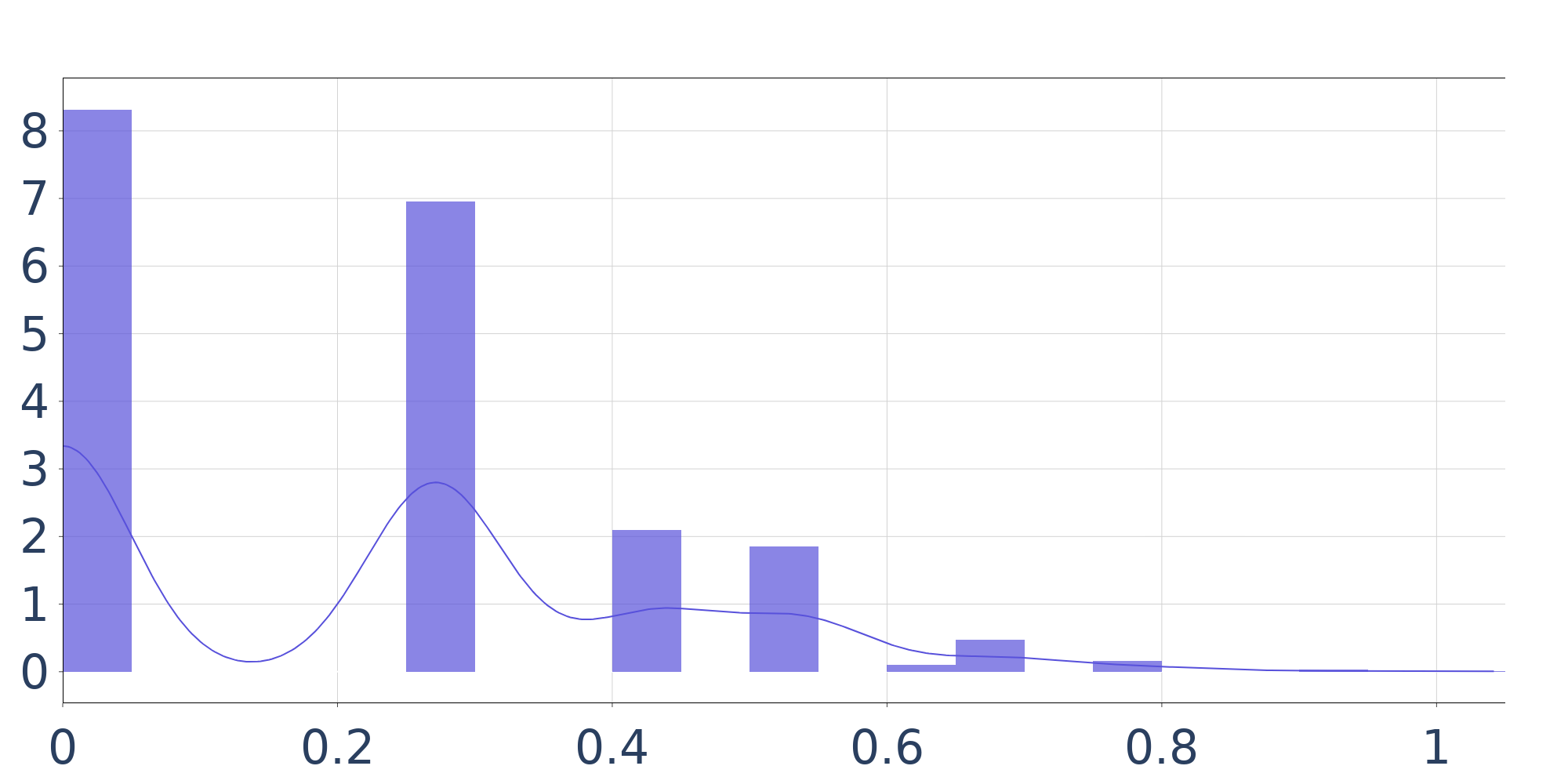}
  \caption{EPR}
\end{subfigure}
\hfill
\begin{subfigure}{\textwidth}
  \centering   \captionsetup{justification=centering} 
  \includegraphics[scale=0.15]{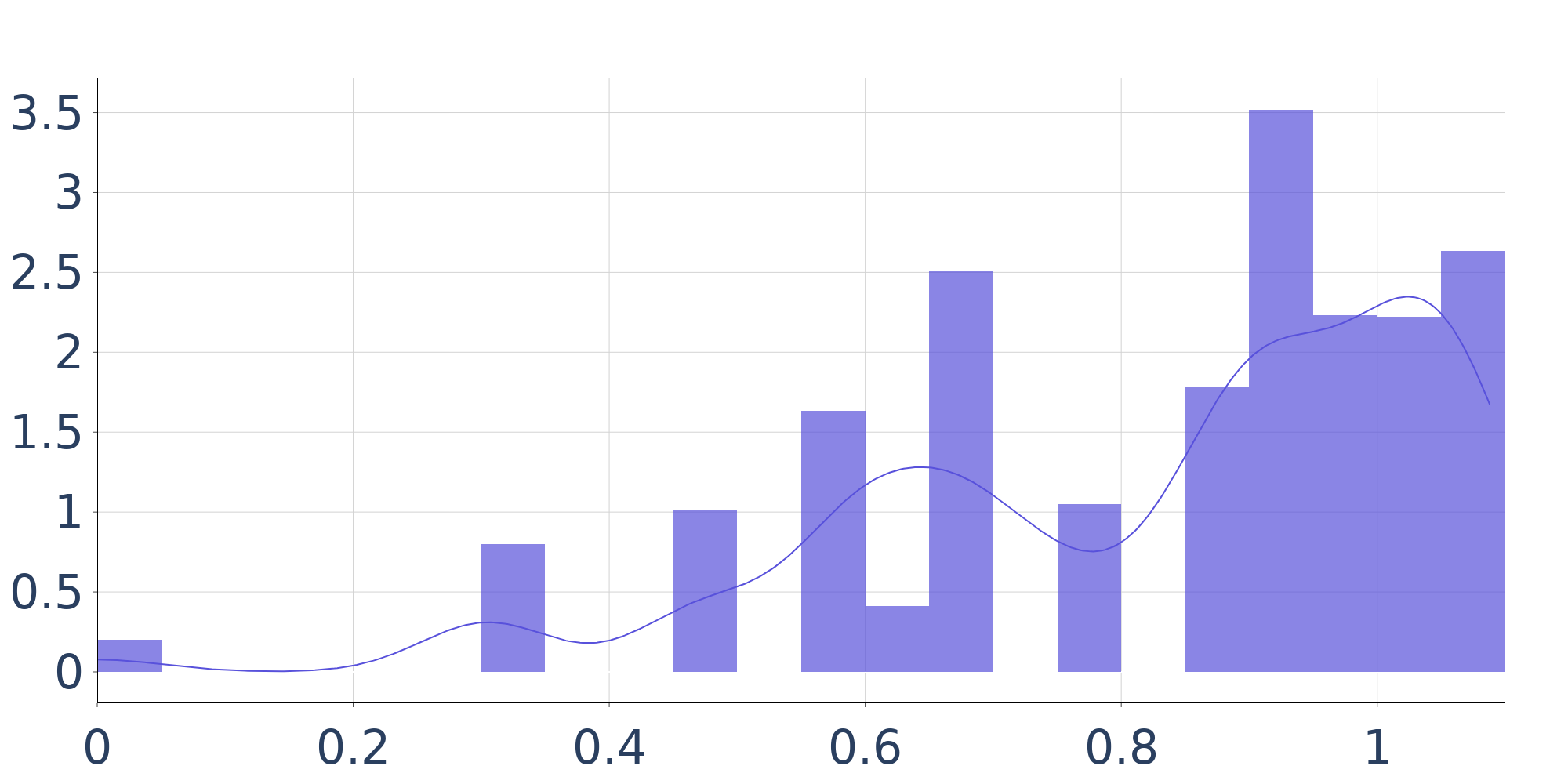}
  \caption{StrategyQA}
\end{subfigure}
\caption{Probability density function of uncertainty estimates of our method using \textbf{GPT3-XL}.}
\label{fig:uncertainty_gpt3xl}
\end{figure*}

\begin{table*}[ht]
\centering
\setlength\tabcolsep{6pt}
\begin{tabular}{c|cc|cc|cc|cc|cc}
\toprule
\multicolumn{1}{c}{} & \multicolumn{2}{c}{GPT3.5} & \multicolumn{2}{|c}{GPT3-XL} & \multicolumn{2}{|c}{GPT4O} & \multicolumn{2}{|c}{Phi3} & \multicolumn{2}{|c}{Mistral} \\
    \cmidrule(lr){2-3}\cmidrule(lr){4-5}\cmidrule(lr){6-7}\cmidrule(lr){8-9}\cmidrule(lr){10-11}
 \multirow{-1}{*}{\textbf{Dataset}} & $\mu$ & $\sigma$ & $\mu$ & $\sigma$ & $\mu$ & $\sigma$ & $\mu$ & $\sigma$ & $\mu$ & $\sigma$ \\
\midrule
GSM8K & 1.21 & 0.53 & 1.55 & 0.48 & 0.30 & 0.35 & 0.45 & 0.48 & 1.28 & 0.76 \\
Fallacy & 0.49 & 0.36 & 0.57 & 0.37 & 0.26 & 0.26 & 0.37 & 0.23 & 0.41 & 0.25 \\
EPR & 0.55 & 0.18 & 0.22 & 0.21 & 0.39 & 0.27 & 0.46 & 0.22 & 0.42 & 0.25 \\
StrategyQA & 0.43 & 0.35 & 0.83 & 0.22 & 0.32 & 0.31 & 0.39 & 0.29 & 0.28 & 0.30 \\
\bottomrule
\end{tabular}
\caption{Mean and standard deviation of uncertainty values as error graph -specific statistics across models.}
\label{tbl:dataset_uncertainty}
\end{table*}

\begin{figure*}[!]
\begin{subfigure}{\textwidth}
\centering   \captionsetup{justification=centering} 
  \includegraphics[scale=0.25]{figures/gsm8k_gpt4o_acc_entropy.png}
  \caption{GSM8K}
\end{subfigure}
\hfill
\begin{subfigure}{\textwidth}
  \centering   \captionsetup{justification=centering} 
  \includegraphics[scale=0.25]{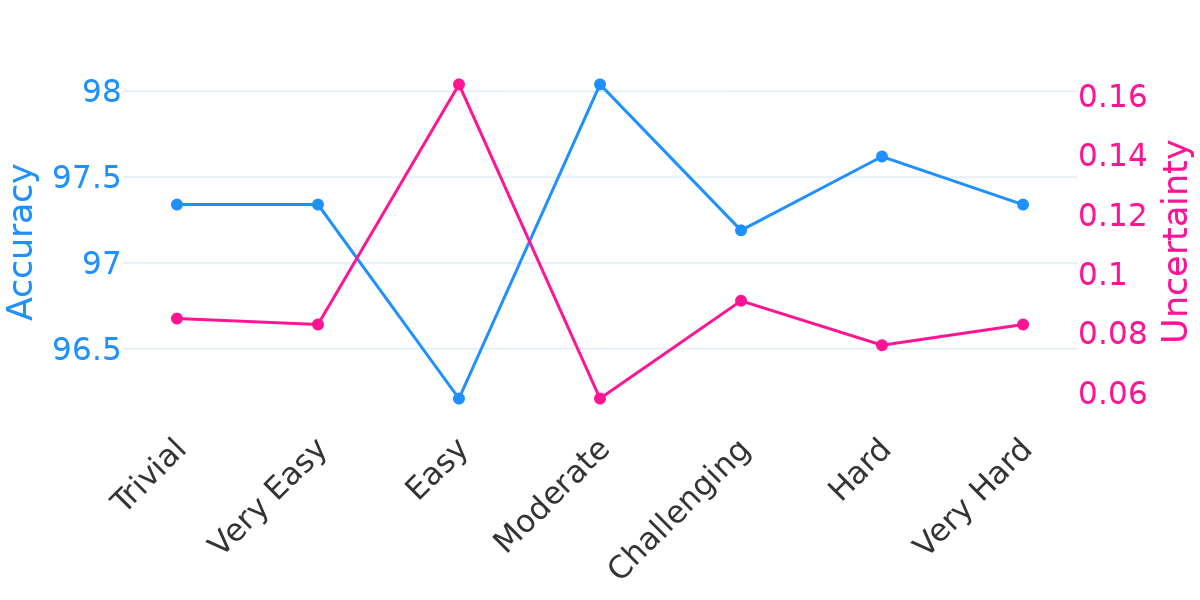}
  \caption{Fallacy}
\end{subfigure}
\hfill
\begin{subfigure}{\textwidth}
  \centering   \captionsetup{justification=centering} 
  \includegraphics[scale=0.25]{figures/strategyqa_gpt4o_acc_entropy.png}
  \caption{StrategyQA}
\end{subfigure}
\hfill
\begin{subfigure}{\textwidth}
  \centering   \captionsetup{justification=centering} 
  \includegraphics[scale=0.25]{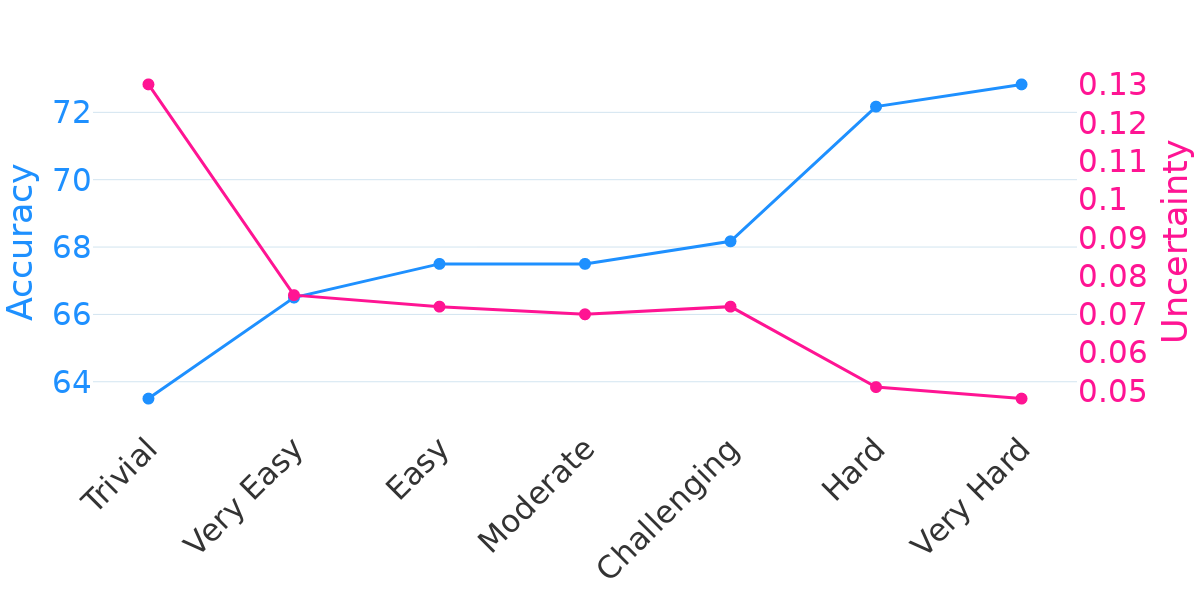}
  \caption{EPR}
\end{subfigure}
\caption{Accuracy vs \textit{Temp-Perb} Uncertainty trend across all selection strategies for\textbf{GPT4o}}
\label{fig:gpt4o_acc_entropy}
\end{figure*}

\begin{figure*}[!]
\begin{subfigure}{\textwidth}
\centering   \captionsetup{justification=centering} 
  \includegraphics[scale=0.25]{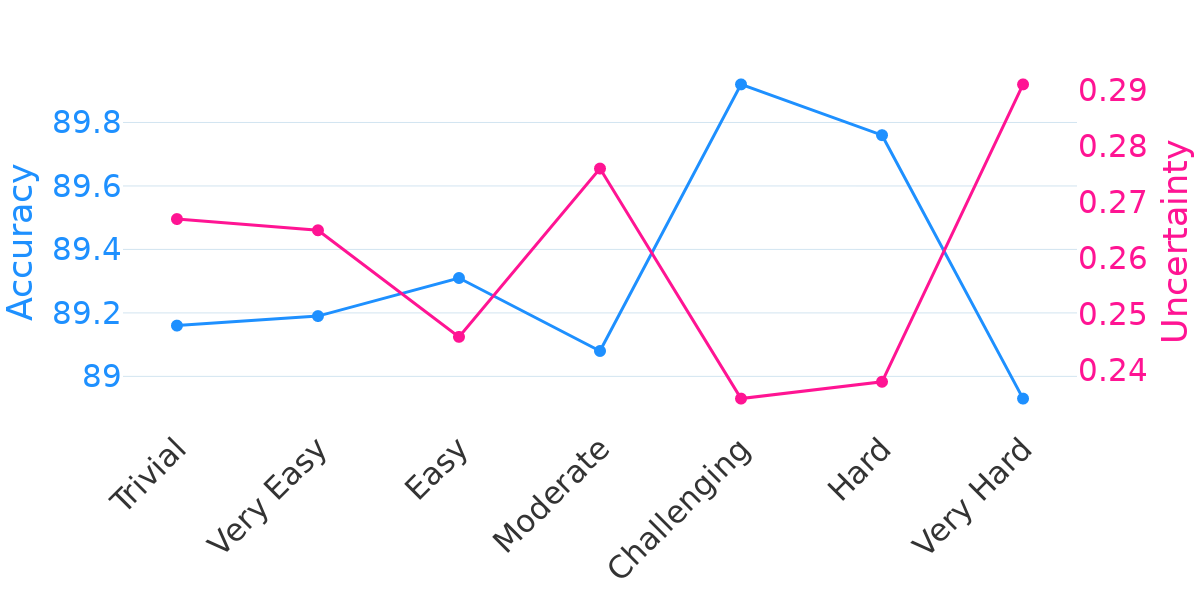}
  \caption{GSM8K}
\end{subfigure}
\hfill
\begin{subfigure}{\textwidth}
  \centering   \captionsetup{justification=centering} 
  \includegraphics[scale=0.25]{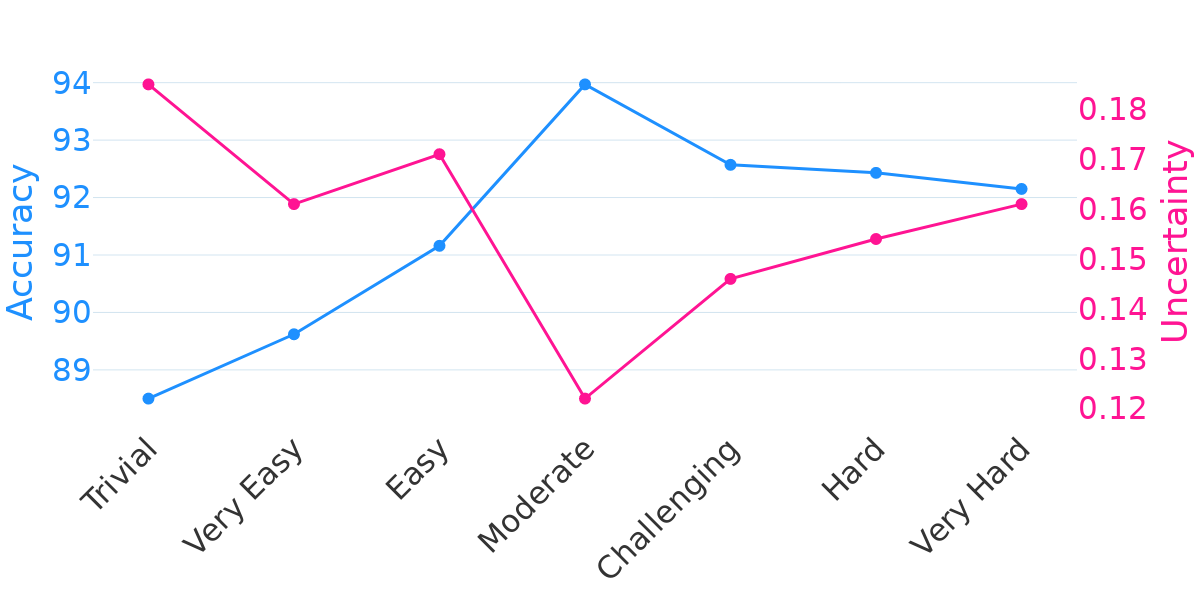}
  \caption{Fallacy}
\end{subfigure}
\hfill
\begin{subfigure}{\textwidth}
  \centering   \captionsetup{justification=centering} 
  \includegraphics[scale=0.25]{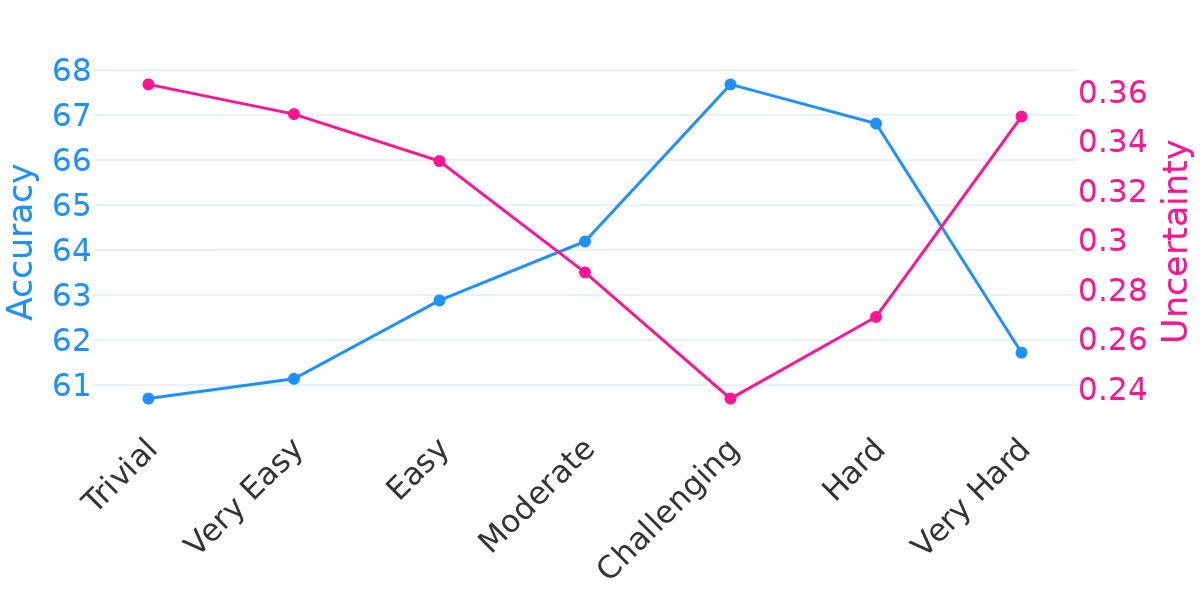}
  \caption{StrategyQA}
\end{subfigure}
\hfill
\begin{subfigure}{\textwidth}
  \centering   \captionsetup{justification=centering} 
  \includegraphics[scale=0.25]{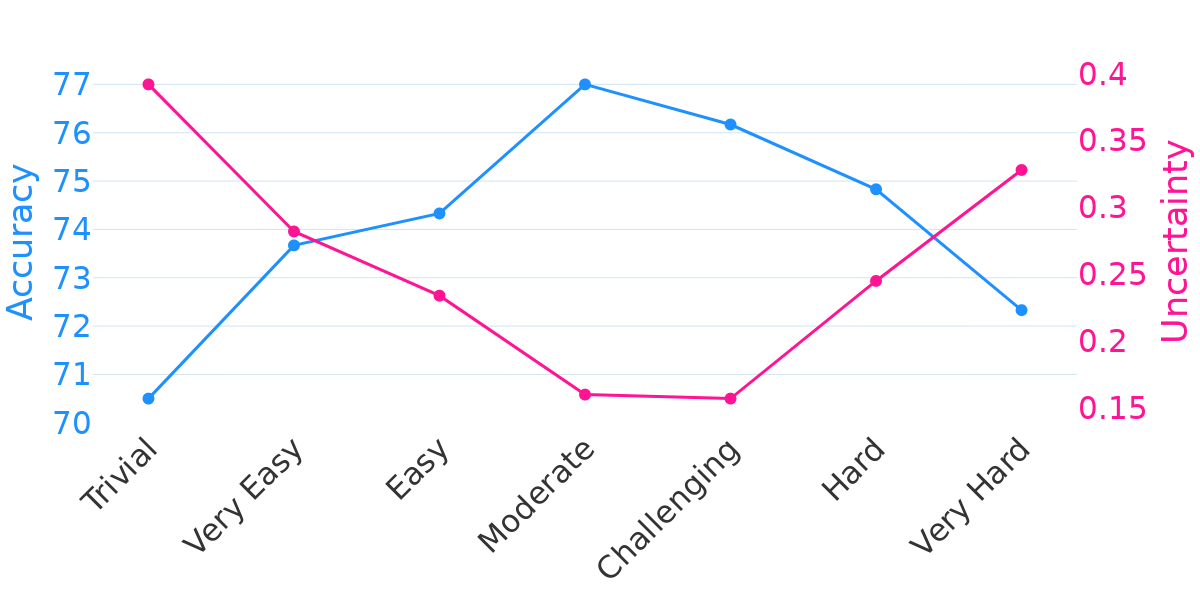}
  \caption{EPR}
\end{subfigure}
\caption{Accuracy vs \textit{Temp-Perb} Uncertainty trend across all selection strategies for\textbf{Phi3}}
\label{fig:phi3_acc_entropy}
\end{figure*}

\begin{figure*}[!]
\begin{subfigure}{\textwidth}
\centering   \captionsetup{justification=centering} 
  \includegraphics[scale=0.25]{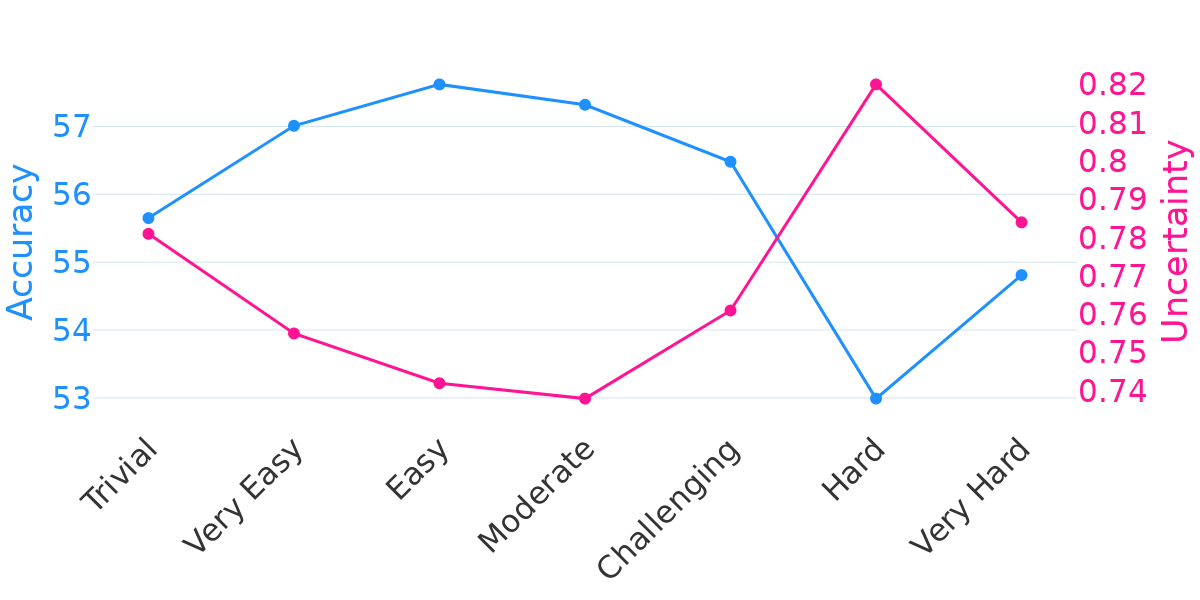}
  \caption{GSM8K}
\end{subfigure}
\hfill
\begin{subfigure}{\textwidth}
  \centering   \captionsetup{justification=centering} 
  \includegraphics[scale=0.25]{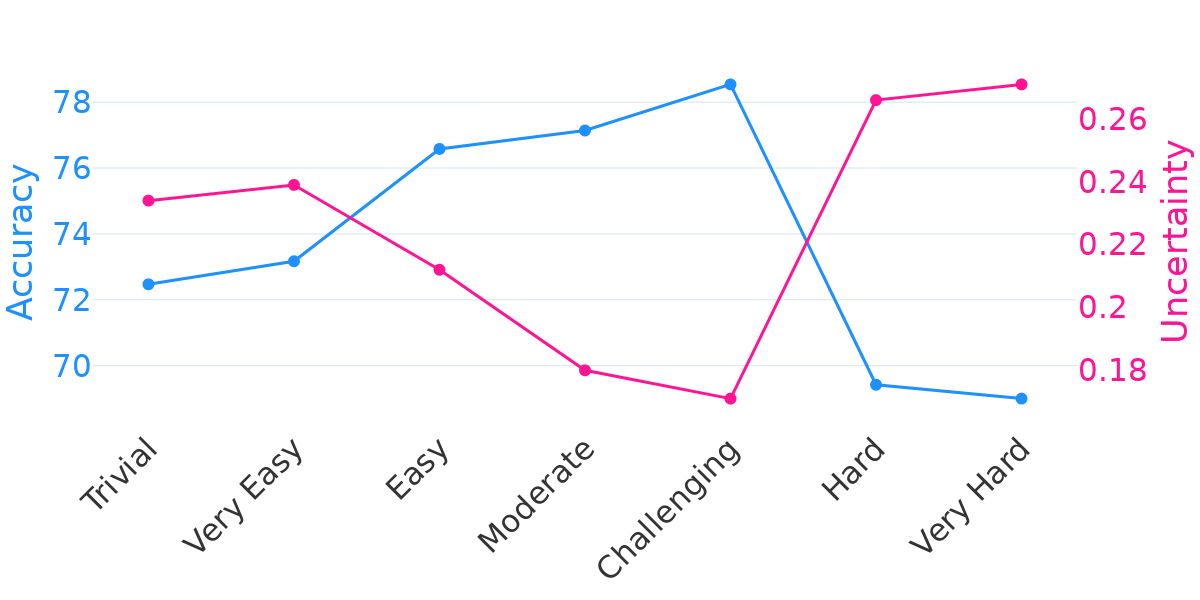}
  \caption{Fallacy}
\end{subfigure}
\hfill
\begin{subfigure}{\textwidth}
  \centering   \captionsetup{justification=centering} 
  \includegraphics[scale=0.25]{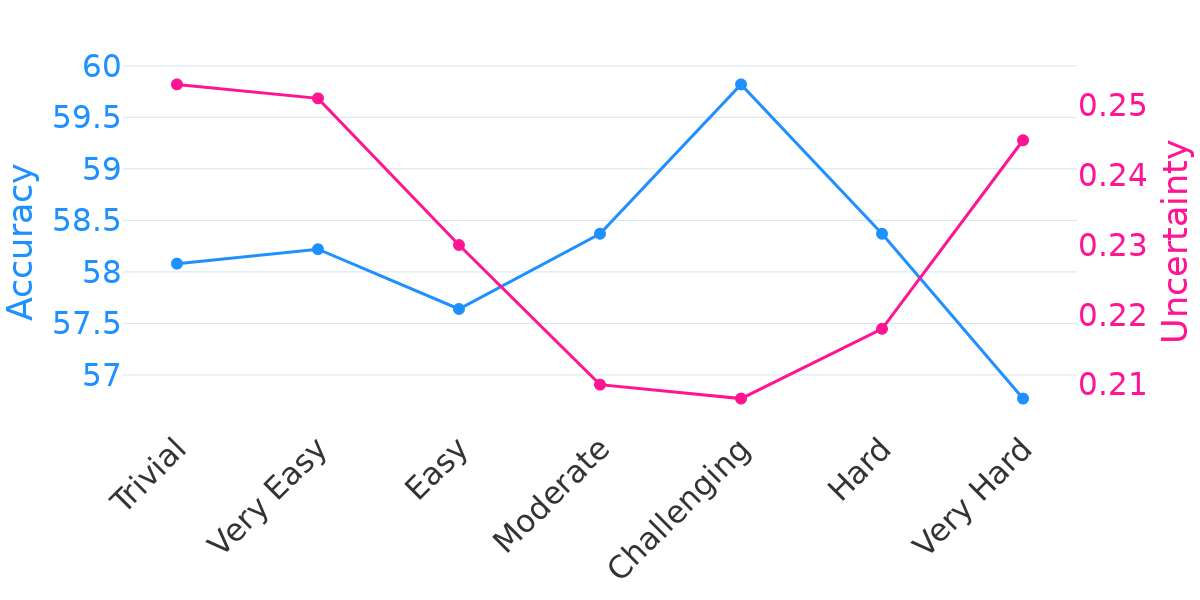}
  \caption{StrategyQA}
\end{subfigure}
\hfill
\begin{subfigure}{\textwidth}
  \centering   \captionsetup{justification=centering} 
  \includegraphics[scale=0.25]{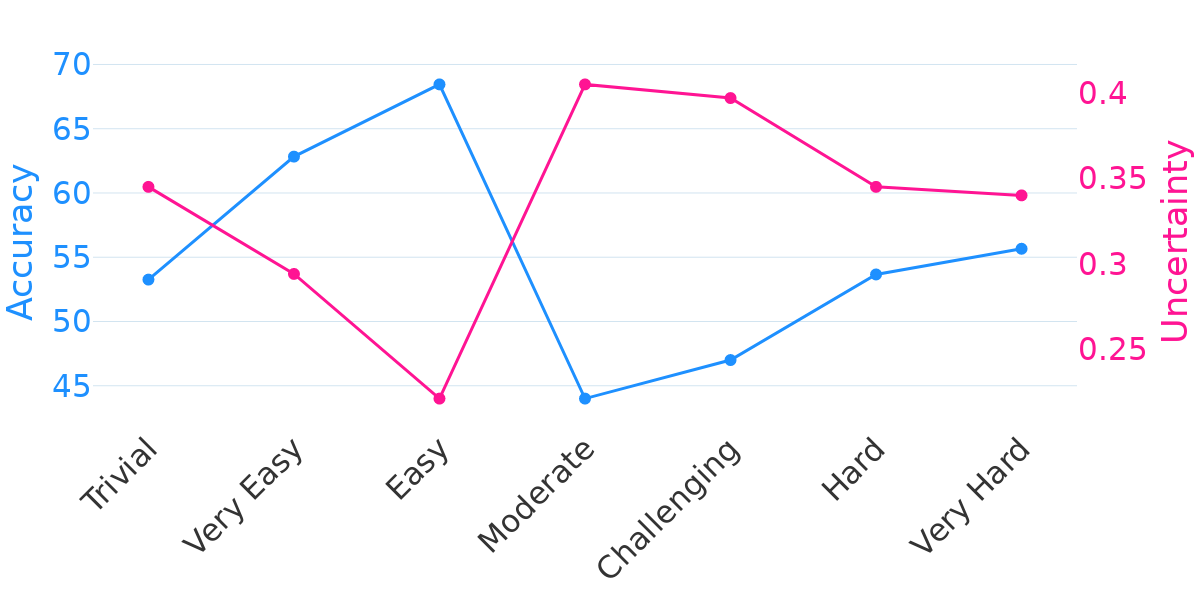}
  \caption{EPR}
\end{subfigure}
\caption{Accuracy vs \textit{Temp-Perb} Uncertainty trend across all selection strategies for\textbf{Mistral}}
\label{fig:mistral_acc_entropy}
\end{figure*}

\begin{figure*}[!]
\begin{subfigure}{\textwidth}
\centering   \captionsetup{justification=centering} 
  \includegraphics[scale=0.25]{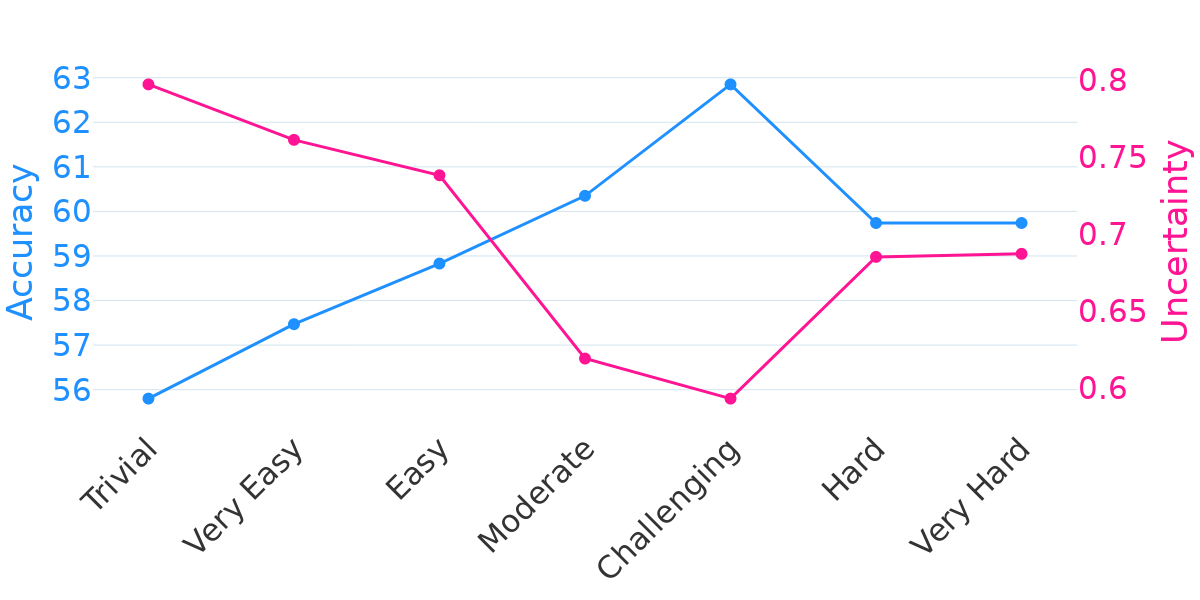}
  \caption{GSM8K}
\end{subfigure}
\hfill
\begin{subfigure}{\textwidth}
  \centering   \captionsetup{justification=centering} 
  \includegraphics[scale=0.25]{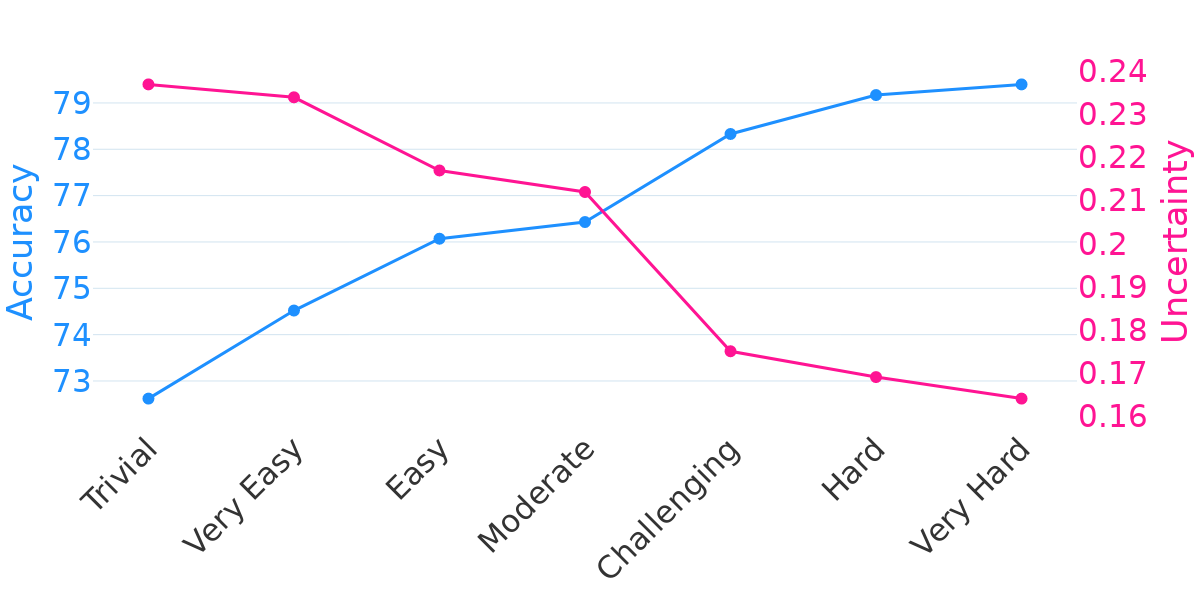}
  \caption{Fallacy}
\end{subfigure}
\hfill
\begin{subfigure}{\textwidth}
  \centering   \captionsetup{justification=centering} 
  \includegraphics[scale=0.25]{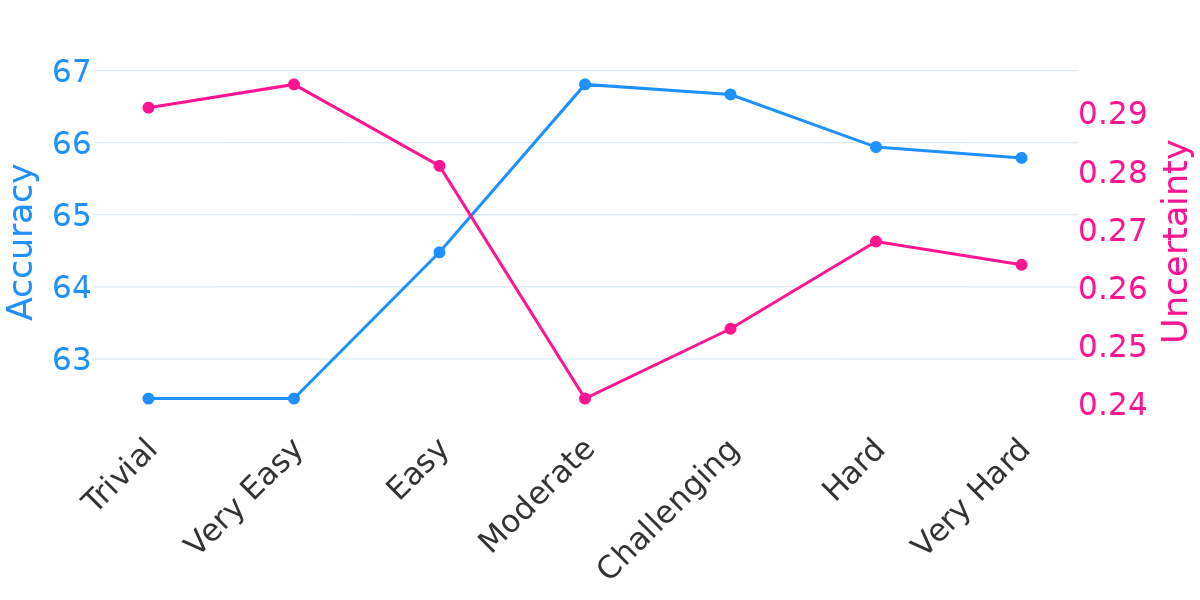}
  \caption{StrategyQA}
\end{subfigure}
\hfill
\begin{subfigure}{\textwidth}
  \centering   \captionsetup{justification=centering} 
  \includegraphics[scale=0.25]{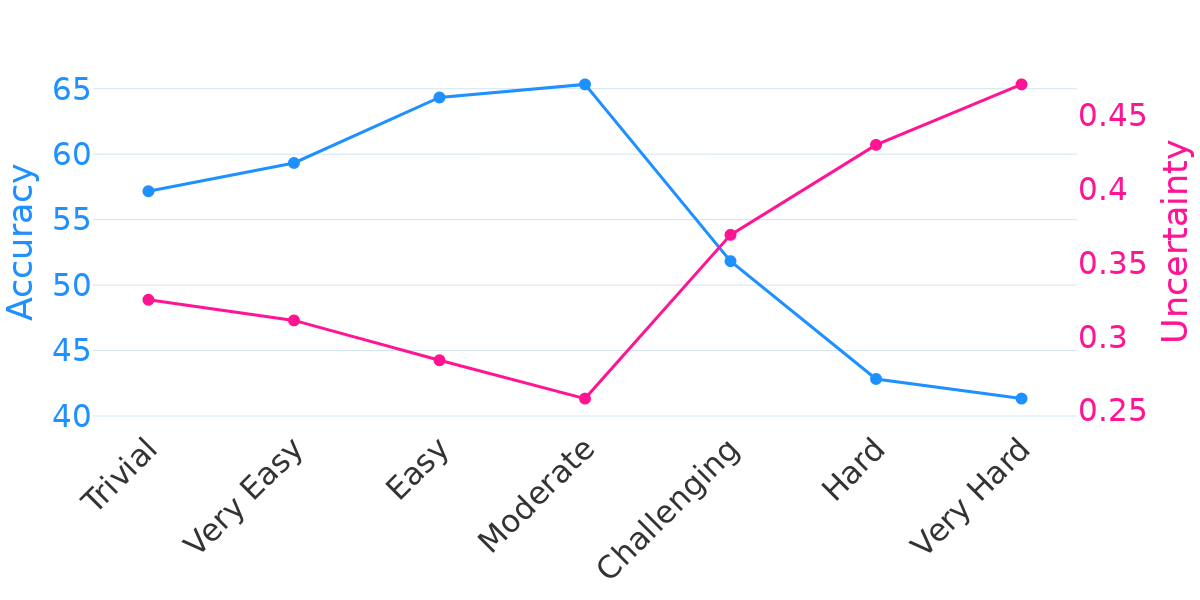}
  \caption{EPR}
\end{subfigure}
\caption{Accuracy vs \textit{Temp-Perb} Uncertainty trend across all selection strategies for\textbf{GPT3.5}}
\label{fig:gpt3.5_acc_entropy}
\end{figure*}

\begin{figure*}[!]
\begin{subfigure}{\textwidth}
\centering   \captionsetup{justification=centering} 
  \includegraphics[scale=0.25]{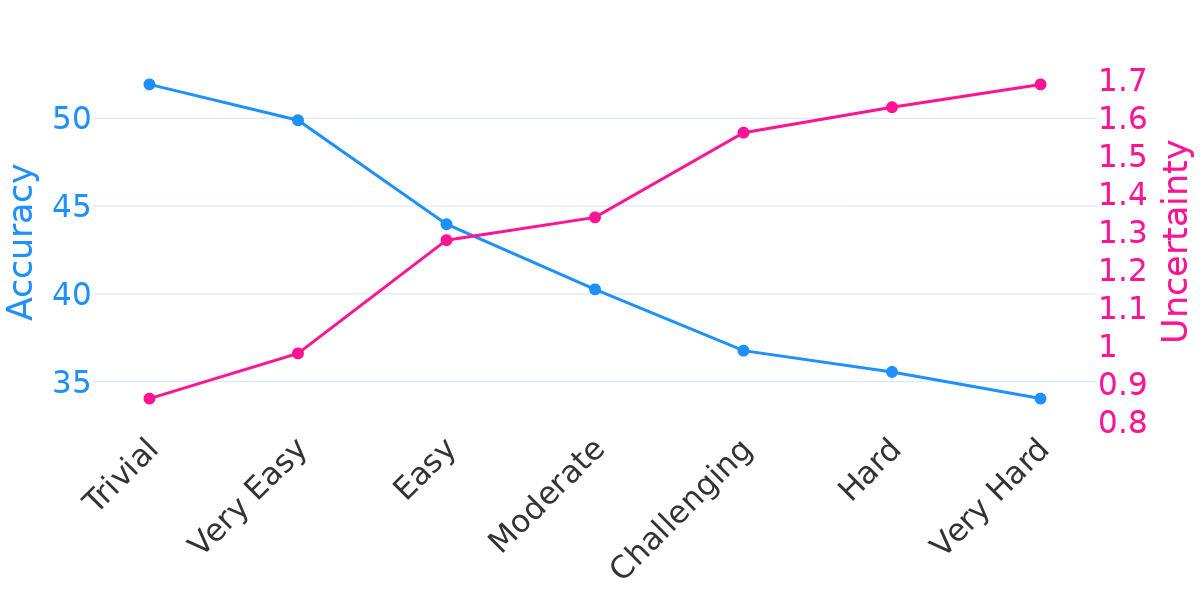}
  \caption{GSM8K}
\end{subfigure}
\hfill
\begin{subfigure}{\textwidth}
  \centering   \captionsetup{justification=centering} 
  \includegraphics[scale=0.25]{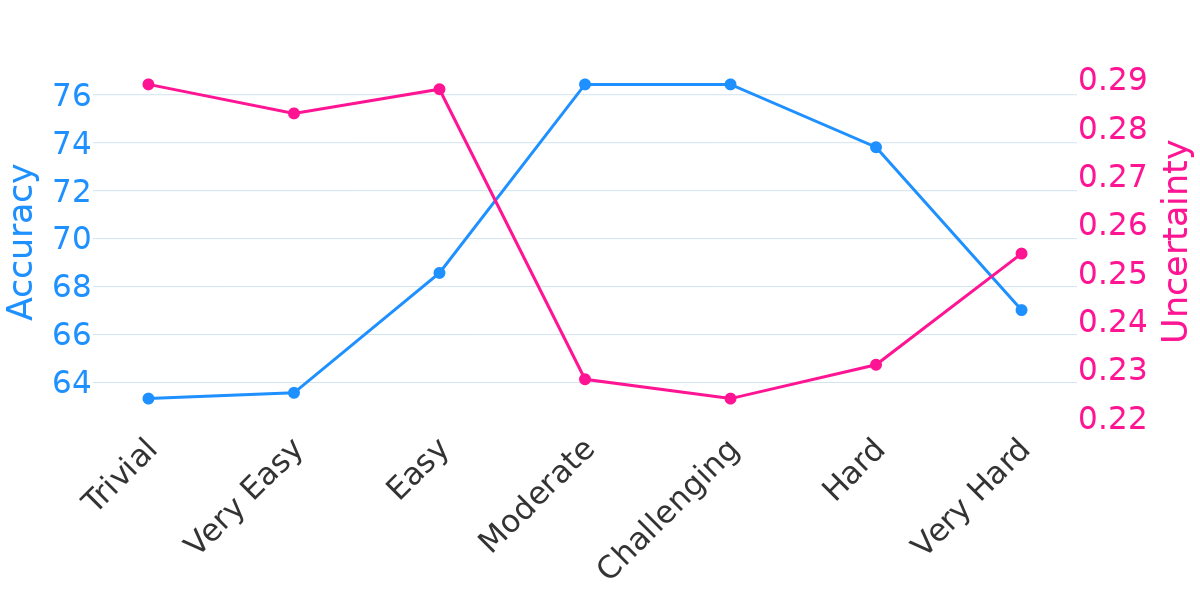}
  \caption{Fallacy}
\end{subfigure}
\hfill
\begin{subfigure}{\textwidth}
  \centering   \captionsetup{justification=centering} 
  \includegraphics[scale=0.25]{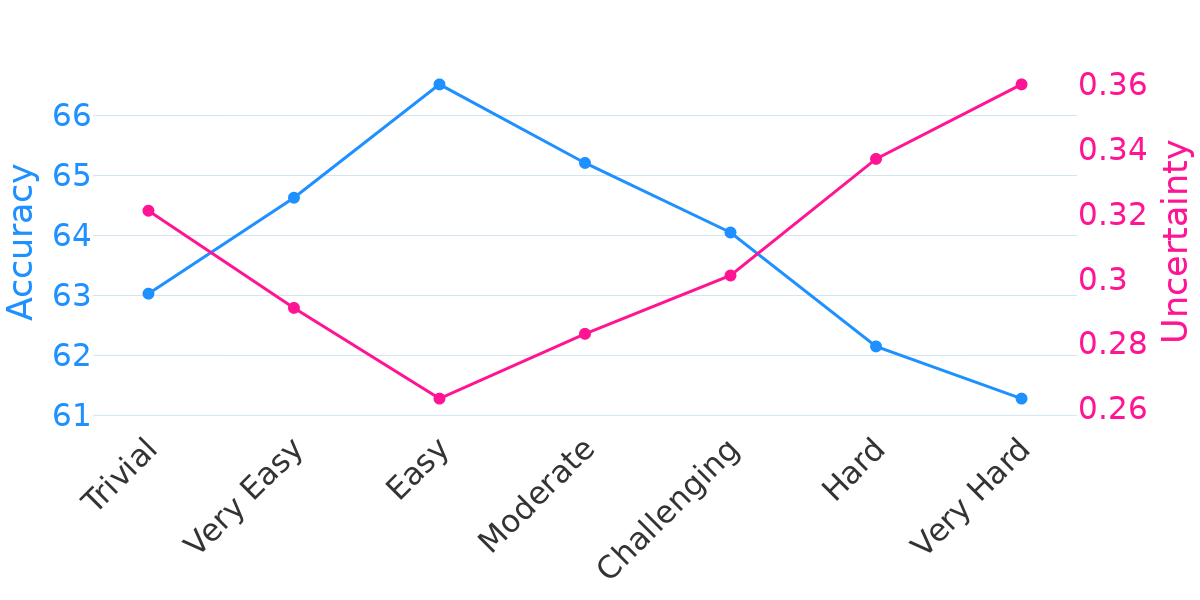}
  \caption{StrategyQA}
\end{subfigure}
\hfill
\begin{subfigure}{\textwidth}
  \centering   \captionsetup{justification=centering} 
  \includegraphics[scale=0.25]{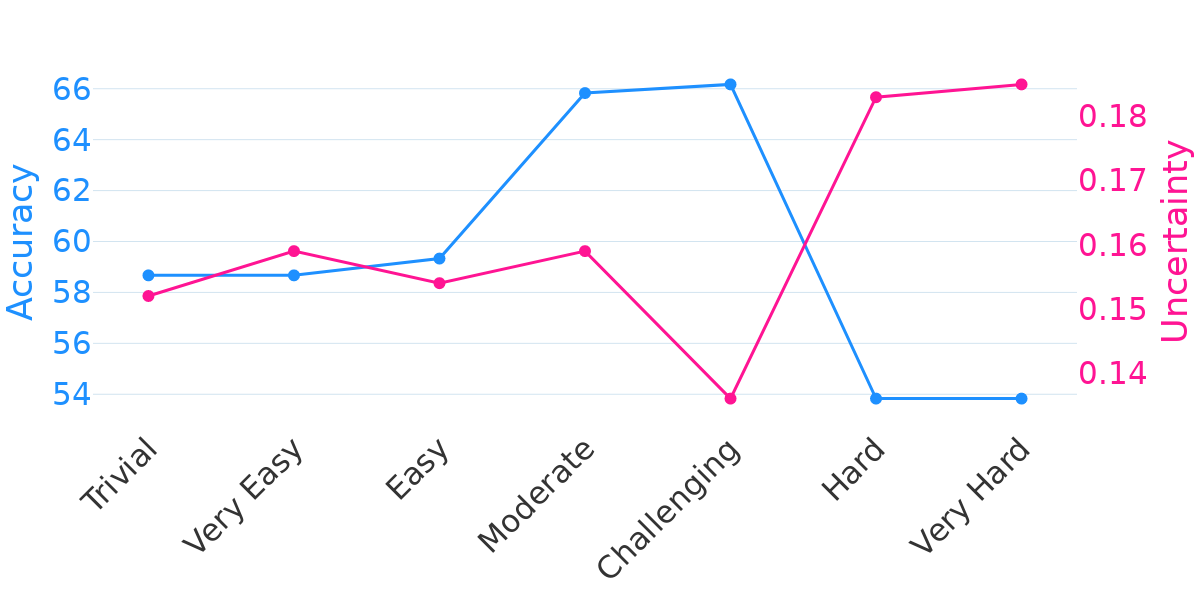}
  \caption{EPR}
\end{subfigure}
\caption{Accuracy vs \textit{Temp-Perb} Uncertainty trend across all selection strategies for\textbf{GPT3-XL}}
\label{fig:gpt3xl_acc_entropy}
\end{figure*}

\end{document}